\documentclass[9.5pt,journal,cspaper,compsoc,final]{IEEEtran}

\let\labelindent\relax
\usepackage{enumitem}
\usepackage{xspace}
\usepackage[nocompress]{cite}
\usepackage{hyperref}
\usepackage[pdftex]{graphicx}
\graphicspath{{./latex/}}

\usepackage{caption}
\usepackage{subcaption}

\usepackage{xcolor}   

\usepackage{amssymb}
\usepackage{soul}

\usepackage[cmex10]{amsmath}
\usepackage{algorithm}
\usepackage{algorithmic}
\usepackage{url}

\usepackage{verbatim}
\usepackage{ifdraft}
\usepackage{color, colortbl}
\definecolor{lightred}{rgb}{0.8,0.36,0.36}

\ifoptionfinal{
\newcommand{\add}[1]{#1}
\newcommand{\del}[1]{{}}
}
{
\newcommand{\add}[1]{{\textcolor{blue}{#1\xspace}}}
\newcommand{\del}[1]{{\textcolor{lightred}{\sout{#1\xspace}}}}
}

\usepackage[normalem]{ulem}

\begin{document}
%
\title{Holistic Measures for Evaluating Prediction Models in Smart Grids}
\label{title}
%
%

\author{Saima~Aman,~\IEEEmembership{Student Member,~IEEE,}
        Yogesh~Simmhan,~\IEEEmembership{Senior Member,~IEEE,}
        and~Viktor~K.~Prasanna,~\IEEEmembership{Fellow,~IEEE}
\IEEEcompsocitemizethanks{\IEEEcompsocthanksitem \copyright 2014 IEEE. Personal use of this material is permitted. Permission from IEEE must be obtained for all other uses, in any current or future media, including reprinting/republishing this material for advertising or promotional purposes, creating new collective works, for resale or redistribution to servers or lists, or reuse of any copyrighted component of this work in other works. 
 
\IEEEcompsocthanksitem S. Aman is with the Department
of Computer Science, University of Southern California, Los Angeles, CA, 90089\protect\\
E-mail: saman@usc.edu
\IEEEcompsocthanksitem Y. Simmhan is with the Indian Institute of Science, Bangalore, India
\IEEEcompsocthanksitem V. Prasanna is with the Department of Electrical Engineering, University of Southern California, Los Angeles, CA, 90089}
\thanks{}}

%
%

\markboth{Transactions on Knowledge and Data Engineering}%
{Aman \MakeLowercase{\textit{et al.}}: Holistic Measures for Evaluating Prediction Models}
%



\IEEEcompsoctitleabstractindextext{
\begin{abstract}
\label{abstract}
The performance of prediction models is often based on ``abstract metrics'' that estimate
the model's ability to limit residual errors between the observed and predicted values. However,
meaningful evaluation \add{and selection} of prediction models for \add{end-user} domains \del{where they are used} requires holistic and
application-sensitive performance measures. Inspired by \add{energy} consumption prediction \add{models} used in the emerging
``big data'' domain of Smart Power Grids, we propose a suite of performance measures to rationally
compare \del{prediction} models along the dimensions of scale independence, reliability, volatility and
cost. We include both application independent and dependent measures, the latter parameterized to
allow customization by domain experts to fit their scenario.  \del{While our measures are generalizable
to other domains, we offer an empirical analysis of our measures using real energy use data
collected over three years to evaluate the goodness of fit of ARIMA time series and regression tree
consumption prediction models for three diverse Smart Grid applications.} \add{While our measures are generalizable
to other domains, we offer an empirical analysis using real energy use data for three Smart Grid applications: planning, customer education and demand response, which are relevant for energy sustainability}. Our results underscore the
value of the proposed measures to offer a deeper insight into models' behavior and their impact on
real applications, which benefit \add{both} data mining researchers and practitioners.
\end{abstract}

}

\maketitle

%

\IEEEdisplaynotcompsoctitleabstractindextext
\IEEEpeerreviewmaketitle

\section{Introduction}
\label{* Sec:introduction}
%
\IEEEPARstart{T}{here} is a heightened emphasis on applying data mining and machine learning techniques to domains
with societal impact, and, in the process, understanding the gaps in existing
research~\cite{wagstaff:icml:2012}. One emerging community contending with a data explosion 
is Smart Power Grids. Advanced instrumentation and controls being deployed to
upgrade aging power grid infrastructure is \del{allowing}\add{offering utilities} unprecedented access to power data at fine
spatial and temporal granularities, \del{and}\add{with} near realtime availability~\footnote{Smart Meter deployments
  continue to rise, US Energy Info. Admin., 2012 \href{http://www.eia.gov/todayinenergy/detail.cfm?id=8590}{www.eia.gov/todayinenergy/detail.cfm?id=8590}}~\cite{simmhan:cise:2013}. \del{However, translating this new-found data to meaningful actions for
ensuring reliable, efficient and sustainable electricity generation and supply requires the novel
use of modeling, mining and evaluation techniques}
\add{However, this data needs to be translated into actionable intelligence by means of novel
  modeling, mining and evaluation methods to
ensure sustainable energy generation and supply}~\cite{ramchurn:cacm:2012}.

One critical opportunity lies in \emph{reliably, accurately and efficiently predicting electric energy consumption for individual premises}. The growing availability of energy consumption data (Kilowatt-Hour (kWh)) from 100,000's of customer smart meters at 15~mins sampling intervals~\cite{shishido:aceee:2012} -- orders of magnitude greater than
traditional readings done once a month -- offers a unique challenge in applying forecasting models
and evaluating their efficacy for Smart Grid applications. The impact of such predictions includes
strategic planning of \add{renewable} generation\del{capacity}, power purchases from energy markets, daily planning to
meet peak power loads, and engaging customers in energy savings programs~\cite{yu:ie:2011,
  mckane:aceee:2008}\add{, all of which can enhance long term energy sustainability and security}. \del{Predictions at
different spatio-temporal granularities (e.g., every 15~min for a residential building, 24~hour for a mall), and horizons (e.g., 8~hours ahead, 1~week ahead) are of use.} 

\add{Historically, data at the feeder and sub-station level have been collected using SCADA
  systems. Hence contemporary} load forecasting models \add{exist}\del{have been proposed} at \add{the} coarser spatial granularity of \del{the} total utility area using Bayesian modeling~\cite{Cottet:JASA:2003}, Support Vector Machines~\cite{chen:powsys:2004}, \add{Artificial Neural Networks} \cite{khotanzad:powsys:1998} and time series methods~\cite{martinez:tkde:2011}\del{since data at the feeder and sub-station level have historically been collected}. However, \add{energy} consumption prediction\footnote{
\add{In this article, we address energy consumption prediction, which deals with average energy over
  an interval (i.e., kWh). This is different from demand (or load) prediction}~\cite{Sugianto:2002:AUPEC, Alfares:ijss:2002} \add{that deals with predicting instantaneous power (kW).}} for individual customers is less studied, both due to the lack of input data and the limited need for such predictions till recently with Smart Grids~\cite{hobby:isgt:2011}. As a result, prediction models at the consumer level, with more intra-day and seasonal variability, and for
innovative Smart Grid applications, \add{which engage customers in sustainability,} have not been well understood nor a \emph{\del{framework} set of performance measures for their
  evaluation} \del{developed} \add{identified}.

The performance of prediction models are often based on ``abstract metrics'', detached from
\add{their} meaningful evaluation for the \add{end-user} domain\del{of their
  use}~\cite{wagstaff:icml:2012}. Common performance measures like Mean Absolute Error (MAE) and
Root Mean Square Error (RMSE) form the basis for selecting a suitable prediction
model. \emph{However, we suggest that these measures are insufficient to evaluate prediction models
  for emerging Smart Grid applications}. \del{The reasons for this go beyond just Smart Grids and are}\add{This gap, discussed
  below, is} relevant to many applied domains \add{beyond just Smart Grids}. 

 \textbf{(1)} The impact of under- and over- predictions can be
 \del{different for}\add{asymmetric on} the domain, and measures
 like RMSE are insensitive to \emph{prediction bias}. For e.g., under-prediction \add{of consumption
   forecasts} is more
 deleterious to Smart Grid applications that respond to peak \del{consumption forecasts}\add{demand}. \textbf{(2)} 
\emph{Scale-dependent} metrics are unsuitable for comparing prediction models applied to different customer sizes. 
\textbf{(3)} The focus on the magnitude of errors overlooks the \emph{frequency} with which
a model outperforms a baseline model or predicts within an error tolerance. Reliable
prediction is key for certain domain applications. \textbf{(4)} \emph{Volatility} is a related
factor that is ignored in common measures, wherein a less volatile prediction model performs
\emph{consistently} better than a baseline model. \textbf{(5)} Lastly, given the ``Big Data''
consequences of emerging applications, the \emph{cost of collecting data and running models} cannot
be disregarded. The extra cost for improved accuracy from a model may be impractical at large
scales in a Smart Grid with millions of customers~\cite{ibm:sg:2012}, or the latency for a
prediction can make it unusable for \del{real-time operations}\add{operational decisions}. These
gaps highlight the need for holistic performance measures to meaningfully evaluate
and compare prediction models \del{for}\add{by domain} practitioners\del{ in application domains}. 

\textbf{Contributions.} 
Specifically, we make the following novel contributions. \del{We identify the need to extend and
introduce measures to evaluate consumption prediction models that go beyond the commonly used error
measures (\S\ref{Sec:related_work}). We classify performance measures along the dimensions
of scale independence, reliability and cost (\S\ref{Sec:measures}), and propose a holistic set of
measures with sound statistical properties to 
assess and rationally compare prediction models. These include six generic measures (\S\ref{Sec:app_ind_measures})
and four application dependent ones (\S\ref{Sec:app_spec_measures}) that can be tuned for scenarios.}\add{(1) We propose a suite of performance
  measures for evaluating prediction models in Smart Grids, defined along three dimensions: scale
  in/dependence, reliability and cost (\S\ref{Sec:measures}). These include two existing measures
  and eight innovative ones (\S\ref{Sec:app_ind_measures}, \S\ref{Sec:app_spec_measures}),
  and also encompass parameterized measures that can be customized for the domain. (2)} We analyze the
usefulness of these concrete measures by evaluating \del{candidate} ARIMA \del{time series} and regression tree
prediction models (\S\ref{Sec:prediction_models}) \del{that are used by}\add{applied to} three Smart Grid applications
(\S\ref{Sec:experimental_setup}) in the Los Angeles Smart Grid Project~\footnote{\href{http://www.smartgrid.gov/project/los_angeles_department_water_and_power_smart_grid_regional_demonstration}{Los Angeles Department of Water and Power: Smart Grid
Regional Demonstration}, US Department of Energy, 2010}. \del{The study uses real consumption data
collected over three years from buildings in the University of Southern California (USC) campus, a
microgrid testbed for the project.} \del{Our analysis of the application independent (\S\ref{Sec:indep_evaluation}) and dependent
measures (\S\ref{Sec:dep_evaluation}) underscores their distinct ability to offer: deeper insight
into models' behavior that can help improve their performance, better understanding of prediction impact on
real applications, intelligent cost-benefit trade-offs between models, and a comprehensive, yet
accessible, goodness of fit for picking the \emph{right} model.
While the measures we introduce are defined in context of Smart Grid Applications, several considerations that went into their selection are relevant for applications in other domains as well.}

\textbf{Significance.} 
\del{Evaluating data mining and machine learning models using meaningful \emph{performance measures} is a key
challenge we address in this article. Our classification of the proposed measures helps estimate
different properties of a model that go beyond just the \emph{magnitude of errors}, and explore bias,
reliability, volatility and cost.} \add{
In this article, we offer meaningful \emph{performance measures} to evaluate predictive
models along dimensions that go beyond just the \emph{magnitude of errors}, and explore bias,
reliability, volatility and cost~\cite{wagstaff:icml:2012}.} \del{For completeness, not}\add{Not} all our measures are novel and some 
extend from other disciplines -- this \add{offers completeness and} also gives a firm statistical grounding.
Our novel application dependent measures with parameterized coefficients set by domain experts allow
apples-to-apples comparison that is meaningful for that scenario \add{\cite{fildes:forecasing:2013}}. A model that seems good using common
error metrics may behave poorly or prove inadequate for a given application; this intuition is validated
by our analysis. All our measures are \del{generalizable to}\add{reusable by} other domains, though they are inspired by and
evaluated for the emerging Smart Grid domain.  

As Smart Grid data becomes widely available, data mining \add{and machine learning}
  research \del{into this under-served domain can reap immense societal benefits}\add{can
provide immense societal benefits to this under-served domain~\cite{ramchurn:cacm:2012}}. Our study based on real
  Smart Grid data \add{collected over
3~years}~\footnote{``Where's the Data in the Big Data
    Wave?'', Gerhard Weikum, SIGMOD Blog, Mar
    2013. \href{http://wp.sigmod.org/?p=786}{wp.sigmod.org/?p=786}} \del{and applications is one
    of}\add{is among} the \emph{first of its kind} in defining holistic measures and
  evaluating candidate consumption models \add{for emerging
microgrid and utility applications}.
\del{Our work offers a common frame of reference for future
researchers and practitioners, and also exposes the gaps in existing predictive research in
addressing this new domain.}
\add{Our analysis of the measures underscores their key ability to offer: deeper insight
into models' behavior that can help improve their performance, better understanding of prediction impact on
real applications, intelligent cost-benefit trade-offs between models, and a comprehensive, yet
accessible, goodness of fit for picking the \emph{right} model.}
\add{Our work offers a common frame of reference for future
researchers and practitioners, while also exposing gaps in existing predictive research for this new domain.}

\section{Related Work}
\label{Sec:related_work}

The performance evaluation of predictive models often involves a single dimension, such as
an error measure, which is simple to interpret and compare\del{ across domains}, but does not necessarily
probe all aspects of a model's performance or its goodness for a given
application~\cite{wagstaff:icml:2012}. A new metric is proposed in ~\cite{fildes:forecasing:2013} based on aggregating performance ratios across time series for fair treatment of over- and under-forecasting. ~\cite{prati:tkde:2011} 
emphasizes the importance of treating predictive performance as a multi-dimensional problem for a more reliable evaluation of the trade-offs between various aspects. \cite{haben:forecasting:2014} introduces a measure to reduce the \textit{double penalty} effect in forecasts whose features are displaced in space or time, compared to point-wise metrics. Further, \cite{Verbraken:tkde:2013} identifies the need for cost-benefit
measures to help capture the performance of a prediction model by a single profit-maximization
metric which can be easily interpreted by practitioners. ~\cite{armstrong:forecasting:1992} highlights the importance of scale, along with
  measures like cost, sensitivity, reliability, understandability, and relationship for decision
  making using forecasts.  Other studies~\cite{sokolova:ai:2006} 
also go beyond standard error measures to include dimensions of sensitivity and specificity. Our effort is
in a similar vein. We propose a holistic set of measures along multiple dimensions to assist
domain users in intelligent model selection for their application, with empirical validation for
emerging Smart Grid domain.

\del{Modeling of demand (or load) is related to, but
different from, electric energy consumption prediction that we consider at the consumer-scale. The former deals with predicting instantaneous power (kW) rather than average energy over an interval (kWh).} 
Existing approaches for consumption prediction include our \add{and other prior} work on regression
tree~\cite{aman:dddm:2011, mori:isgt:2011}, \del{and}
time series models~\cite{amjady:powsys:2001}, \del{and} artificial neural networks and expert
systems~\cite{martinez:tkde:2011, Kramer:has:2010, khotanzad:powsys:1998}. In practice, utilities 
use simpler \del{methods such as averaging consumption of previous several days}\add{averaging
  models based on recent consumption}~\cite{nyiso-model,
coughlin:lbnl:2008}. In this spirit, our baseline models for comparative evaluation consider Time of
the Week (ToW) and Day of the Week (DoW) averages.

As \del{in other disciplines}\add{elsewhere}, Smart Grid literature \del{commonly}\add{often} evaluates
predictive model performance in terms of the magnitude of errors between the observed and predicted
values~\cite{jewell:itec:2012,chakraborty:aaai:2012}. \del{The common}\add{Common} statistical
measures \del{, well-defined in}\add{for}
time series forecasting~\cite{armstrong:forecasting:1992} are \del{the} \emph{Mean Absolute Error (MAE)},
the \emph{Root Mean Square Error (RMSE)} and the \emph{Mean Absolute Percent Error (MAPE)}, the latter given by: 
\begin{equation} \label{eq:MAPE}
MAPE = \frac{1}{n} \sum_{i=1}^{n}\frac{|p_{i} - o_{i}|}{o_{i}}
\end{equation}
where $o_{i}$ is the observed value at interval $i$, $p_{i}$ is the model predicted value, and
$n$ is the number of intervals for which the predictions are made.
\del{A variant of MAPE, }\emph{Mean Error Relative (MER)} to the mean of the observed, has also been used 
to avoid the effects of observed values close to zero~\cite{martinez:tkde:2011}. \del{Sometimes, the }RMSE
values normalized by the mean of observed values, called the \emph{coefficient
  of variation of the RMSE (CVRMSE)}, is \add{also} used~\cite{dong:energynbuildings:2005, aman:dddm:2011}:
\begin{equation} \label{eq:CVRMSE}
CVRMSE = \frac{1}{\overline{o}} \sqrt{ \frac{1}{n} \sum_{i=1}^{n}(p_{i} - o_{i})^{2}}
\end{equation}
where $\overline{o}$ is the mean of the $n$ observed values, and $p_i, o_i$ and $n$ are as before. 
While these measures offer a necessary statistical standard of model performance,
they by themselves are inadequate due to the reasons listed before, viz., their inability to address
prediction-bias, reliability, scale independence and cost of building models and making
predictions. 

Some researchers have proposed application-specific metrics. ~\cite{mathieu:energynbuildings:2011} defines metrics related to Demand-Response. 
The Demand Shed Variability Metric (SVM) and Peak Demand Variability Metric (PVM) help \del{in reducing}\add{reduce} over-fitting and extrapolation errors that increase error variance or introduce prediction bias. 
Our application-\emph{dependent} (rather than -\emph{specific}) measures are defined more broadly,
with measure \emph{parameters} that can be tuned for diverse applications that span even beyond
Smart Grids. 

Relative measures help compare a prediction model with a baseline model~\cite{armstrong:forecasting:1992}. \emph{Percent Better} gives the
fraction \add{of} forecasts by a model that are more accurate than a random walk model.  This is a unit-free measure that is also immune to outliers present in the series, by discarding information about the amount of change. 
The \emph{Relative Absolute Error (RAE)}, calculated as a ratio of forecast error for a model to the corresponding error for the random walk, is simple to interpret and communicate to the domain users.
The prediction horizon also has an impact on model performance. 
\emph{Cumulative RAE}~\cite{armstrong:forecasting:1992} is defined as the ratio of the arithmetic sum of the absolute error for the proposed model over the forecast horizon and the corresponding error for the random walk model. Relative and horizon metrics have been used less often for smart grid prediction models.

\del{Hyndman and Koehler~\cite{hyndman:forecasting:2006} categorize prediction }\add{Prediction} error
metrics \add{have been categorized} into scale-dependent measures, percentage errors, relative errors and scaled errors~\cite{hyndman:forecasting:2006}. RMSE and MAE are \emph{scale-dependent} \del{that are}\add{and} applicable to \del{data sets}\add{datasets} with similar \del{scales}\add{values}. \add{Scale-independent} \emph{Percentage errors} like MAPE and CVRMSE \del{are scale-independent and} can \del{thus} be used to compare performance on \del{data sets}\add{datasets} with different magnitudes. \emph{Relative measures} are determined by dividing model errors by the errors of a baseline model, while \emph{scaled errors} remove the scale of data by normalizing the errors with the errors obtained from a baseline prediction method.

In summary, standard statistical measures of performance for predictive models may not be
\add{adequate or} meaningful
for domain-specific applications, while narrowly defined measures for a single application are not
reusable or comparable across applications. This gap is particularly felt in the novel
domain of Smart Grids. Our work is an attempt to address this deficiency by introducing a suite of
performance measures along several dimensions, while also leveraging existing measures where appropriate.

\section{Performance Measure Dimensions}
\label{Sec:measures}

Performance measures that complement standard statistical error measures for evaluating prediction models fall along several dimensions that we discuss here.

\textbf{Application In/dependent}:
Application independent measures are specified without knowledge of how predictions from the model are
being used. These do not have any specific dependencies on the usage scenario and can be used as a
uniform measure of comparison across different candidate models. Application dependent measures incorporate
\emph{parameters} that are determined by specific usage scenarios of the prediction model. The measure
formulation itself is generic but requires users to set values of parameters (e.g., acceptable error
thresholds) for the application. These allow a \del{more} nuanced evaluation of prediction models
\del{suited}\add{that is customized} for the application in concern.

\textbf{Scale-Independent Errors}:
In defining error metrics, the residual errors are measured as a difference between the observed and predicted values. So, if $o_{i}$ is the $i^{th}$ observed value and $p_{i}$ is the $i^{th}$
predicted value, then the \emph{scale-dependent} residual or prediction error, $e_{i} = p_{i} - o_{i}$.
MAE and RMSE 
are based on residual errors, and suffer from
being highly dependent on the range of observed values. Scale-independent errors, on the other hand, are usually normalized against the observed value and hence better suited for comparing model
performance across different magnitudes of observations. 

\textbf{Reliability}:
Reliability offers an estimate of how consistently a model produces similar results. This dimension is important to understand how well a model will perform on a yet unseen data that the system will encounter in future, relative to the data used while testing. A more reliable model provides its
users \add{with} more confidence in its use. Most commonly used measures fail to consider the
frequency of \add{acceptable model} performance over a period of time, which we address through measures we introduce.

\textbf{Cost}:
Developing a prediction model has a cost associated with it in terms of effort and time for data
collection, training and using the models. The number of times a trained model can be reused is also
a factor. Data cost is \del{an}\add{a particularly} important consideration in this age of ``Big Data'' since quality
checking, maintaining and storing large feature-sets can be \del{unwieldy}\add{untenable}. Compute costs can even be intractable when prediction models are used millions of times within short periods. 

\vspace{-0.05in}
\section{Application Independent Measures}
\label{Sec:app_ind_measures}
\del{This section discusses application independent measures. }Several standard statistical measures
with well understood theoretical properties fall in \del{this}\add{the} category \add{of application
  independent measures}. For completeness, we recognize two relevant, \del{ones}\add{existing
  scale-independent measures}, MAPE~\cite{jewell:itec:2012,martinez:tkde:2011} and CVRMSE~\cite{dong:energynbuildings:2005, aman:dddm:2011}\del{, as providing scale-independent measures}. More importantly, we
introduce novel application independent measures, along the reliability and cost dimensions, and their properties. 

\vspace{-0.05in}
\subsection{\textbf{Mean Absolute Percentage Error (MAPE)}}
This is a variant of MAE, normalized by the observed value\footnote{\add{MAPE is not defined if
    there are zero values in the input, which is rare as energy consumption (kwh) values are
    generally non-zero due to always present base consumption (unless there is a black-out), and can
    be ensured by data pre-processing.}} at each interval (\ref{eq:MAPE}), thus
providing \textit{scale independence}. \del{This is a simple but well understood error measure that
  is expressed as a percentage.} \add{It is simple to interpret and commonly used for evaluating
  predictions in energy and related domains}~\cite{martinez:tkde:2011, jewell:itec:2012,
  hyndman:2007, chen:powsys:2004, chakraborty:aaai:2012}.

\vspace{-0.05in}
\del{{\textbf{Coefficient of Variation of the Root Mean Square Error (CVRMSE)}}}
\add{\subsection{\textbf{Coefficient of Variation of RMSE (CVRMSE)}}}
It is the normalized version of the \del{commonly used}\add{common} RMSE \add{measure}, \del{where the RMSE is
divided by} \add{that divides it by} the average \footnote{\add{CVRMSE is not defined if this average is zero, which is rare
    as energy consumption (kwh) values are generally positive (unless there is net-metering), and
    can be ensured by data pre-processing.}}  of the observed values (\ref{eq:CVRMSE}) to offer
\textit{scale independence}. This \del{measure} is an unbiased estimator that incorporates both the
\add{prediction model} bias \del{of the prediction model as well as}\add{and} its variance, and
gives a \add{unit-less} percentage error measure\del{ that is unitless}. CVRMSE \del{tends to be}\add{is} sensitive to infrequent large errors due to the squared term.

\vspace{-0.05in}
\subsection{\textbf{Relative Improvement (RIM)}}
We propose RIM as a \textit{relative measure} for \textit{reliability} that is estimated as the
frequency of predictions \del{from}\add{by} a candidate model \add{that are} better than a baseline
model.
RIM is a simple, unitless measure that complements error measures \del{when}\add{in cases
  where} being \add{accurate} more often \del{accurate}than a baseline is useful, and occasional large errors relative to the baseline are acceptable.

\begin{equation} \label{eq:RIM}
RIM = \frac{1}{n} \sum_{i=1}^{n} C(p_{i},o_{i},b_{i})
\end{equation}
where $o_i, p_i$ and $b_i$ are the observed, model predicted and baseline predicted values for interval $i$, and $C(p_{i},o_{i})$ is a count function defined as:
\begin{equation}
C(p_{i},o_{i},b_{i}) = \begin{cases} 1, & \mbox{if } |p_{i} - o_{i}| < |b_{i} - o_{i}|  \\ 0, & \mbox{if } |p_{i} - o_{i}| = |b_{i} - o_{i}| \\ -1, & \mbox{if } |p_{i} - o_{i}| > |b_{i} - o_{i}| \end{cases}
\end{equation}

\subsection{\textbf{Volatility Adjusted Benefit (VAB)}}
VAB is another measure for \textit{reliability} that captures how consistently a candidate model outperforms a baseline model by normalizing the model's error improvements over the baseline by the standard deviation of these improvements. Inspired by the \emph{sharpe ratio}, this \textit{relative measure} offers a ``risk
adjusted'' scale-independent error value. The numerator captures the relative improvement of the candidate model's MAPE over the baseline's (the benefit). If these error improvements\footnote{\add{The error improvements offered by a given model over the baseline model are expected to have normal distribution for VAB to be meaningful.}} are consistent
across $i$, then their standard deviation would be low (the volatility) and the VAB high. But, with
high volatility, the benefits would reduce reflecting \add{a} lack of consistent improvements.
\begin{equation} \label{eq:VAB}
VAB = \frac{\frac{1}{n}\sum_{i=1}^{n} (\frac{|b_i - o_i|}{o_{i}} - \frac{|p_i - o_i|}{o_{i}})}{\sigma(\frac{|b_i - o_i|}{o_{i}} - \frac{|p_i - o_i|}{o_{i}})}
\end{equation}
where $o_i, p_i$ and $b_i$ are the observed, model predicted and baseline predicted values for interval $i$.

\subsection{\textbf{Computation Cost (CC)}}
\label{Sec:app_ind_measures:cost}
The \textit{cost} for training and predicting using a model can prove important when it is used either at large scales and/or in realtime applications that are sensitive to prediction latency.  CC is defined in seconds as the sum of the wallclock time required to train a model, $CC_{t}$, and the wallclock time required to make predictions using the model, $CC_{p}$, for a given prediction duration with a certain horizon.  Thus, $CC = CC_{t} + CC_{p}$.

\subsection{\textbf{Data collection Cost (CD)}}
Rather than examine the raw size of data used for training or predicting using a model, a more useful measure is the effort required to acquire and assemble the data. Size can be managed through cheap storage but collecting the necessary data often requires human and organizational effort. We propose a \textit{scale-dependent} measure of data \textit{cost} defined in terms of the number of \emph{unique values} of features involved in a prediction model. CD is defined for a particular
training and prediction \emph{duration} as the sum of $n_{s}$, the number of static (time-invariant) features that require a one-time collection effort, and $n_{d}$, the number of dynamic features that need periodic acquisition.
\pagebreak[0]
\begin{equation} \label{eq:CD}
CD = \sum_{i=1}^{n_{s}} [s_{i}] + \sum_{i=1}^{n_{d}} [d_{i}]
\end{equation}
where $[s_{i}]$ and $[d_{i}]$ are the counts of the unique values for the feature $s_{i}$ and
$d_{i}$ respectively.	

\vspace{-0.1in}
\section{Application Dependent Measures}
\label{Sec:app_spec_measures}

Unlike the previous measures, application dependent performance measures are \emph{parameterized} to suit
specific usage scenarios and can be customized by domain experts to fit their needs. The novel
measures we propose here are themselves not narrowly defined for a single application (though they
are motivated by the needs observed in the smart grid domain). Rather, they are generalized through the use of
coefficients that are themselves application specific. We group them along the dimensions
that we introduced earlier.

\vspace{-0.05in}
\subsection{\textbf{Domain Bias Percentage Error (DBPE)}} 
We propose DBPE as a signed \textit{percentage error} measure that offers \textit{scale
  independence}\del{ and}\add{. It} indicates if the predictions are
positively or negatively biased compared to the observed values\del{. This is an important
  consideration}\add{, which is important}
when \del{errors with a particular bias (i.e., }over- or under-prediction \add{errors}, relative to
observed, \del{)} have
\del{different considerations}\add{a non-uniform impact on the application}. We define DBPE as an asymmetric loss function based on the sign bias. \del{The}\add{Granger's}
linlin function\del{introduced by Granger }~\cite{granger:or:1969} is suitable for this \del{purpose} as it is linear on both
sides of the origin but with different slopes on each side. The asymmetric slopes allow different
penalties for positive/negative errors.
\begin{equation} \label{eq:DBPE}
DBPE = \frac{1}{n}\sum_{i=1}^{n}\frac{\mathcal{L}(p_{i},o_{i})}{o_{i}}
\end{equation}
where $\mathcal{L}(p_{i},o_{i})$ is the linlin loss function defined as:
\begin{equation}
\mathcal{L}(p_{i},o_{i}) = \begin{cases} \alpha \cdot |p_{i}-o_{i}|, & \mbox{if } p_{i} > o_{i}  \\  0, & \mbox{if } p_{i} = o_{i}  \\ \beta \cdot |p_{i}-o_{i}|, & \mbox{if } p_{i} < o_{i}  \end{cases}
\end{equation}
where $o_i$ and $p_i$ are the observed and model predicted values for the interval $i$, and $\alpha$
and $\beta$ are penalty parameters associated with over- and under- prediction,
respectively. $\alpha$ and $\beta$ are configured for specific application and the ratio
$\alpha/\beta$ measures the relative cost of over-prediction to under-prediction for that
application~\cite{Torgo:DS:2009}. Further, we introduce a constraint that $\alpha + \beta = 2$ to
provide DBPE the interesting property of reducing to MAPE when $\alpha=\beta=1$. 
\vspace{-0.05in}
\subsection{\textbf{Reliability Threshold Estimate (REL)}}
Often, applications may care less about the absolute error\add{s} \del{values} of a model's predictions and
\del{instead} prefer an estimate of how frequently the errors fall within a set threshold \add{that the
  application can withstand}. We define REL as
the frequency of prediction errors that are less than an application determined error threshold, $e_t$. 
\begin{equation} \label{eq:REL}
REL = \frac{1}{n} \sum_{i=1}^{n} C(p_{i},o_{i})
\end{equation}
where $o_i$ and $p_i$ are the observed and \add{the} model predicted values for the interval $i$, and $C(p_{i},o_{i})$ is a count function defined as:
\begin{equation}
C(p_{i},o_{i}) = \begin{cases} 1, & \mbox{if } \frac{|p_{i} - o_{i}|}{o_{i}} < e_{t}  \\ 0, & \mbox{if } \frac{|p_{i} - o_{i}|}{o_{i}} = e_{t} \\ -1, & \mbox{if } \frac{|p_{i} - o_{i}|}{o_{i}} > e_{t} \end{cases}
\end{equation}
\subsection{\textbf{Total Compute Cost (TCC)}}
In the context of an application, it is meaningful to supplement the data and compute costs (CD and
CC) with an estimate of the total running cost of using a model \emph{for a duration of interest}
specific to that application. We define the parameters:
\begin{itemize}
\item
$\tau$, the number of times a model is trained within the duration,
\item
$\pi$, the number of times a model \del{is used to} make\add{s} predictions with a given \del{prediction} horizon, in that duration.
\end{itemize}
These parameters are not just application specific but also vary by the candidate model, based on
how frequently it needs to be trained and its effective prediction horizon. 
We define the total training cost in seconds for a prediction duration based on $\tau$ and $\pi$, and the unit
costs for training and prediction using the model, $CC_{t}$ and $CC_{p}$, introduced in \S~\ref{Sec:app_ind_measures:cost}:
\begin{equation} \label{eq:TCC}
TCC = CC_{t} \cdot \tau + CC_{p} \cdot \pi
\end{equation}
\subsection{\textbf{Cost-Benefit Measure (CBM)}} Rather than treat cost in a vacuum, it is
worthwhile to consider the cost for a model relative to the gains it provides. CBM \del{is useful in
comparing}\add{compares} candidate models having different error measures and costs to evaluate which provide\add{s} a
high reward for \add{a} unit compute cost spent. 
\begin{equation} \label{eq:CBM}
CBM = \frac{(1-DBPE)}{TCC}
\end{equation}
The numerator \del{offers}\add{is} an estimate of the accuracy (one minus error measure) while the denominator is
the compute cost. \del{While} We use DBPE as the error measure and TCC as the cost, \add{but} these can be\del{suitably}
replaced by other \add{application dependent} error measures (e.g., CVRMSE, MAPE) and costs (e.g.,
$CD$, $CC_p$). A model \del{that offers a}\add{with} high accuracy but \del{is prohibitively
  costly}\add{prohibitive cost} may \del{not be
  suitable for an application}\add{be unsuitable}. 


\section{Candidate Prediction Models}
\label{Sec:prediction_models}
\del{In our evaluation of the proposed measures using Smart Grid applications as exemplars, we use two
candidate models for performing consumption predictions, discussed in brief here. These models were
chosen based on our own prior study~\cite{aman:dddm:2011} as well as existing literature discussed
in \S~\ref{Sec:related_work}.}
\add{The candidate models for evaluation of the proposed measures were selected based on our \del{own} prior study~\cite{aman:dddm:2011} as well as existing literature discussed in \S~\ref{Sec:related_work}.}

\textbf{Time Series Model}:
\label{time_series_model}
A time series \add{(TS)} model predicts the future values of a variable based on its
previous observations. The ARIMA (Autoregressive Integrated
Moving Average) model is a commonly used \del{time series}\add{TS} prediction model. 
\del{It is defined in terms of three parameters: $p, d,$ and $q$, where $d$ is the number of times a time series needs to be differenced to make it stationary; $p$ is the autoregressive order that captures the number of past values; and $q$ is the moving average order that captures the number of past white noise error terms.}
\add{It is defined in terms of the number ($d$) of times a time series needs to be differenced to make it stationary; the autoregressive order ($p$) that captures the number of past values; and the moving average order ($q$) that captures the number of past white noise error terms.}
These parameters are determined using autocorrelation and partial autocorrelation functions, using the Box-Jenkins test~\cite{jenkins:1970}. 

ARIMA is simple to use as it does not require knowledge of the underlying
domain~\cite{amjady:powsys:2001}. However, estimating the model parameters, $d, p,$ and $q$, requires human examination of the partial correlogram of the time series, though some automated functions\del{that} perform a partial parameter sweep to \del{pick}\add{select} these values.\del{ are available.} 

\textbf{Regression Tree Model}:
\label{regression_tree_model}
A regression tree \add{(RT)} model~\cite{breiman:cart:1984} is a kind of decision tree that recursively partitions the data space into
smaller regions, until a constant value or a linear regression model can be fit for the smallest
partition. \del{Each partition is a cut along one feature of a multi-dimensional feature space, and each
node in the tree represents a conditional decision on one of these \del{feature's}\add{features'} value, with the leaf
node providing the regression function.}

Our earlier work on \del{a regression tree}\add{an RT} model for campus microgrid consumption prediction identified
several advantages~\cite{aman:dddm:2011}. It's flowchart style tree structure helps interpret the impact of different features on consumption. \del{This allows us to prioritize feature selection
while assisting domain users to interpret the results.} Making predictions on a trained model is fast
though collecting feature data and training the model can be costly. \del{The time ordering of data
is not relevant since every interval is identified by a feature vector.} It can be used to make
predictions far into the future if the feature values are available. \del{It is also resilient to missing
values for a feature by taking the average of all the leaf node values in the last reachable
sub-tree.}

\section{Experimental Setup}
\label{Sec:experimental_setup}

We validate the efficacy of our proposed performance measures \del{ under real-world}\add{for real world} \del{application
conditions}\add{applications}. The USC campus
microgrid~\cite{simmhan:buildsys:2011} is a testbed for the DOE-sponsored Los Angeles Smart Grid
Project. \del{The} ARIMA and Regression Tree prediction models are
used \del{within this project} to predict energy consumption at 24-hour and 15-min granularities\add{,} for the
entire campus and for 35 individual buildings. \del{In this study}\add{Here}, we consider the campus and four
representative buildings: \emph{DPT}, \del{housing} a small department with teaching and office space;
\emph{RES}, a \del{large} suite of residential dormitories with decentralized control of cooling and appliance power
loads; \emph{OFF}, hosting administrative offices and telepresence lab\del{ space}; and \emph{ACD}, a
large academic teaching building. These buildings were considered after several pilot studies to
provide diversity in terms of floor size, age, end use, types of occupants, and net electricity
consumption.

\subsection{Datasets}
\textbf{Electricity Consumption Data~\footnote{The electric consumption
  datasets used in this article are available upon request for academic use.}:} We used 15-min granularity electricity consumption data \del{for
the campus and the four buildings} collected by the USC Facility Management Services
between 2008 to 2010 (Table~\ref{table:dataset}). These gave $3 \times 365 \times 96$ or $\sim$100K
samples per building. We linearly interpolated missing values ($<$3\% of samples) and aggregated\del{the 96 intervals of} 15-min data in each day to get the 24-hour granularity values ($\sim$1K samples
per building). Observations from 
2008 and 2009 were used for training the models while the
predictions were evaluated against the \add{out-of-sample} observed values for 2010~\footnote{\add{The 24-hour data was available
only till Nov 2010 at the time of experiments, and hence 24-hour models are tested for a 11 month
period. The 15-min models span the entire 12 months.}}. \del{(Note: The 24-hour data was available
only till Nov 2010 at the time of experiments, and hence 24-hour models are tested for a 11 month
period. The 15-min models span the entire 12 months).}

\textbf{Weather Data~\footnote{NOAA Quality Controlled Local Climatological Data,
    \href{http://cdo.ncdc.noaa.gov/qclcd/}{cdo.ncdc.noaa.gov/qclcd/}}:} We collected historical
hourly average and maximum temperature data \del{from}\add{curated by}
NOAA for\del{the} Los Angeles/USC Campus\del{region} for 
2008--2010. These values were linearly interpolated to get 15-min values. We also collected daily
maximum temperatures that were used for the 24-hour granularity models.

\textbf{Schedule Data~\footnote{USC Academic Calendar, \href{http://academics.usc.edu/calendar/}{academics.usc.edu/calendar/}}:}
We gathered campus information related to the semester periods, working days and holidays from USC's
Academic Calendar.

\begin{table}[!t]
\renewcommand{\arraystretch}{1.0}
\caption{\textbf{Electricity consumption dataset.} Summary statistics of the campus microgrid
  consumption data for training years 2008-2009, and testing year 2010, at different spatial and
  temporal granularities.  
}
\centering  
\begin{tabular}{l r r r r} 
\hline\hline                  
Entity & \multicolumn{2}{c}{Mean \emph{(kWh)}} & \multicolumn{2}{c}{Std. Deviation \emph{(kWh)}} \\
& Training & Testing & Training & Testing\\  
\hline
\textbf{Campus} & &&&\\
24-hour data & 462,970 & 440,803 & 52,956 & 43,454\\
15-min data  & 4,823 & 4,377 & 809 & 770 \\
\hline
\textbf{DPT} & & &&\\
24-hour data & 405.64 & 405.56  & 112.39 & 108.14\\
15-min data & 4.23 & 4.16 & 1.93 & 1.98\\
\hline
\textbf{RES} & & &&\\
24-hour data & 4,220.30 & 3,670.56  & 1,809.00 & 1,460.08\\
15-min data & 43.97 & 37.79  & 22.36 & 17.93\\
\hline
\textbf{OFF} & & &&\\
24-hour data &2,938.90 & 2,790.70 & 591.97 & 549.37\\
15-min data &30.66 & 28.42 & 13.05 & 10.03\\
\hline
\textbf{ACD} & & &&\\
24-hour data & 4,466.40 & 4,055.85 & 640.92 & 552.64\\
15-min data & 46.65& 41.30 & 14.08 & 13.09\\
\hline
\end{tabular}
\label{table:dataset} 
\end{table}

\subsection{Model Configurations}
\label{config}

  \textbf{Regression Tree \add{(RT)} Models:}
For 24-hour \add{(granularity)} predictions\del{ (also called 24-hour predictions)}, we used five features for the
\del{regression
tree}\add{RT} model: Day of the Week (Sun-Sat), Semester (Fall, Spring, Summer), Maximum and
Average Temperatures, and a Holiday/Working day flag. For the 15-min \add{(granularity)}
predictions\del{ (also
called 15-min predictions)}, we used five features: Day of the Week, Time of Day (1--96,
representing the 15-min slots in a day), Semester, \del{Maximum }temperature, and Holiday/Working day
flag.\del{ Temperatures were numerical features while the rest were categorical features.} The \del{regression
tree}\add{RT} model was trained \add{once} using \emph{MATLAB's} \texttt{classregtree} function~\cite{breiman:cart:1984} to
find an optimally pruned tree. 
\del{We refer to the regression tree models as
\textbf{RT} in the analysis.}

  \textbf{ARIMA Time Series \add{(TS)} Models:}
For 24-hour predictions, the ARIMA models \add{are \emph{retrained} and used to} make predictions every week for four different prediction
horizons: 1-week, 2-week, 3-week, and 4-week ahead. Unlike RT, the performance of time series models differ
by the prediction horizon. We use a moving window over the past 2~years for
training these models with $(p,d,q) = (7,1,7)$, equivalent to a 7~day lag\del{. These parameters
were statically set}\add{, selected} after examining several variations. For 15-min
predictions, we \add{\emph{retrain} models and }predict every 2~hours for three different horizons: 2-hour,
6-hour, and 24-hour ahead. We use a moving window over the past 8~weeks for training, with $(p,d,q)
= (8,1,8)$, equivalent to a 2~hour lag. We used the \texttt{arima} function in the \emph{R forecast
  package}~\cite{hyndman:2007} for constructing the time series models.  This function used
conditional sum of squares (CSS) as the fitting method. 
 \del{We refer to the ARIMA time series model as \textbf{TS} in the following.}

  \textbf{Baseline Models:}
For 24-hour predictions, we selected the Day of
Week mean (DoW) as the baseline, defined for each day of the week as the kWh value for that day
averaged over the training period (i.e., 7 values from averaging over 2008 \del{\&}\add{and} 2009). DoW was chosen
over Day of Year (DoY) and Annual Means since it consistently out-performed \del{the others}\add{them}.
For 15-min predictions, we selected the Time of the Week mean (ToW) as the baseline, defined for
each 15-min in a week as \add{the} kWh value for that interval averaged over the training period (i.e., $7
\times 96$ values from averaging over 2008 \del{\&}\add{and} 2009). Here too, ToW \del{was picked as it was
significantly better than the}\add{out-performed} Time of the Year (ToY) and Annual Means. 

\subsection{Smart Grid Applications}
\label{Sec:use_cases}
We \del{now} introduce three applications, \del{in use}\add{used} within the USC \del{campus} microgrid, to evaluate our
proposed measures. \del{The exemplars ground the evaluation in real-world benefits while not detracting
from the generalized use of our measures.}

  \textbf{Planning:}
Planning capital infrastructure such as building remodeling and power system upgrades for energy
efficiency trades-off investment against electric power savings. Medium to long term electricity
consumption predictions at coarse (24-hour) granularity for campus and individual buildings help
this decision making. Such models are run six times in a year.

  \textbf{Customer Education:}
Educating power customers \del{about}\add{on} their energy usage can enhance their participation in \add{energy sustainability} by curtailing
demand and meet\add{ing} monthly budgets~\cite{ramchurn:cacm:2012}. One form of education
\del{occurs}\add{is} through \add{giving}
consumption \del{predictions provided}\add{forecasts} to customers in a building \del{through}\add{on} \del{online}\add{web} and mobile
apps~\footnote{USC SmartGrid Portal,
  \href{http://smartgrid.usc.edu}{smartgrid.usc.edu}}. Building-level predictions at\del{ both} 24-hour and
15-min granularities are made during the day\del{time} (6AM-10PM).

  \textbf{Demand Response:}
Demand Response (DR) optimization is a critical technique \add{for achieving energy sustainability}
enabled by smart grids. \del{where}\add{In DR,} customers are
encouraged to \del{reduce}\add{curtail} consumption, \emph{on-demand}, to reduce \add{the chance of}
black-outs when peak power usage periods are
anticipated by the utility~\cite{mathieu:energynbuildings:2011}. Historically, these high peaks occur
between 1-5PM on weekdays~\footnote{DWP TOU Pricing,
  \href{http://bp.ladwp.com/energycredit/energycredit.htm}{bp.ladwp.com/energycredit/ energycredit.htm}}, and
predictions during these periods over the 
short time horizon at 15-min granularity are vital for utilities to decide when to initiate curtailment requests from
customers or change their pricing. Often, the predictions are before, at the beginning of, and
during the high peak period.
\section{Analysis of Independent Measures}
\label{Sec:indep_evaluation}

We first examine the use and value of the six application independent measures
(\S~\ref{Sec:app_ind_measures}) to \del{broadly} evaluate the candidate models for predicting
\add{campus and building} consumption\del{ of the campus and four buildings} at coarse and fine \del{temporal}\add{time} granularities.

\vspace{-0.05in}
\subsection{24-hour Campus Predictions}
\begin{figure*}[!t]
\begin{minipage}[b]{0.33\textwidth}
\centering
  \begin{subfigure}[b]{\textwidth}
\centering
   \includegraphics[width=\textwidth] {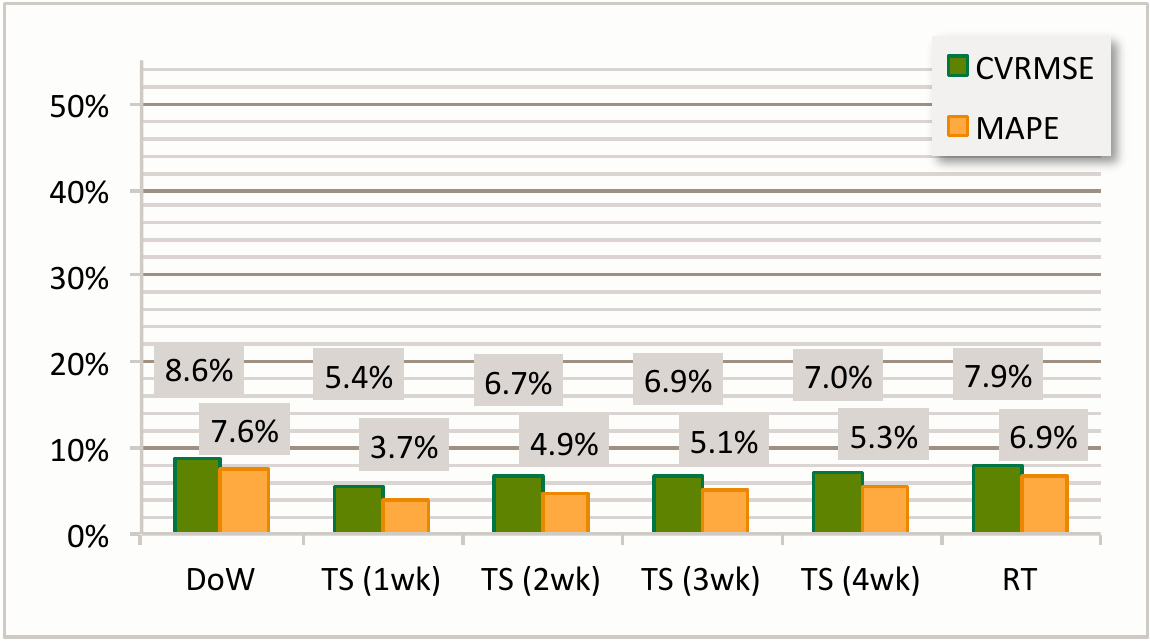}
   \caption{Campus}
	\label{fig:C24-CV-MAPE}
 \end{subfigure}
 \begin{subfigure}[b]{\textwidth}
   \includegraphics[width=\textwidth] {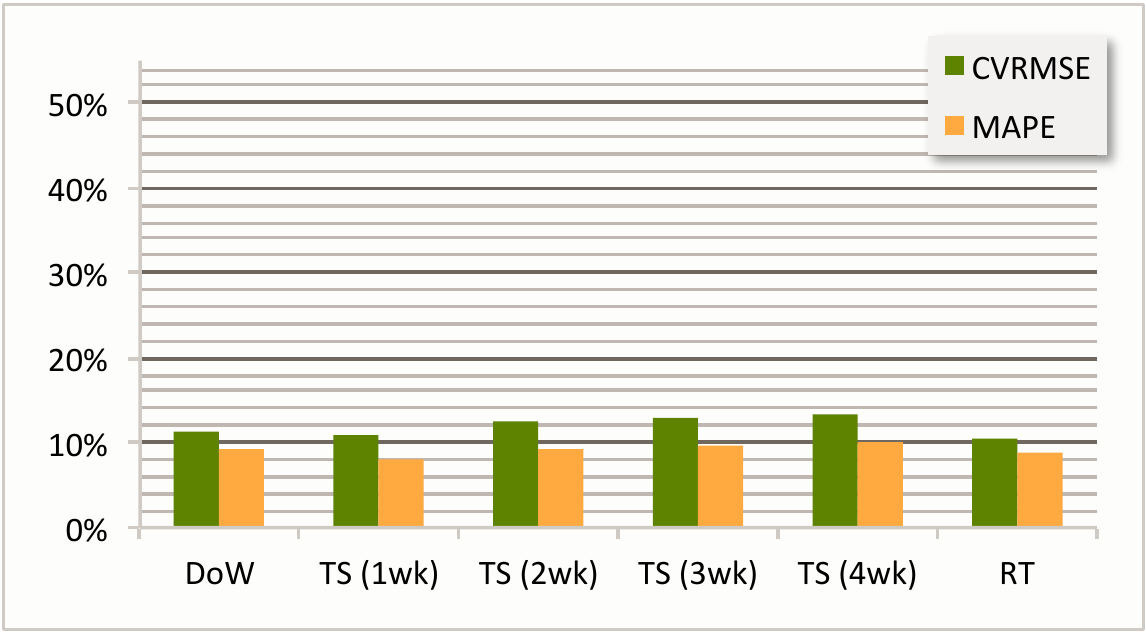}
   \caption{DPT}   
	\label{fig:MHP24-CV-MAPE}
 \end{subfigure}
 \begin{subfigure}[b]{\textwidth}
   \includegraphics[width=\textwidth] {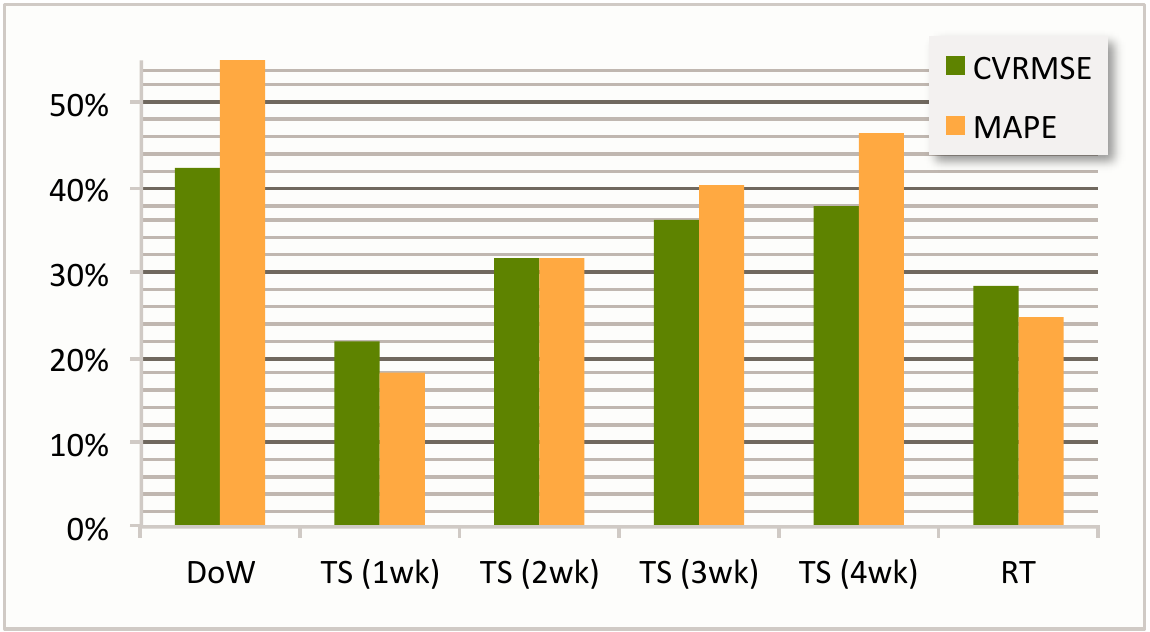}
   \caption{RES}   
	\label{fig:FLT24-CV-MAPE}
 \end{subfigure}
 \begin{subfigure}[b]{\textwidth}
   \includegraphics[width=\textwidth] {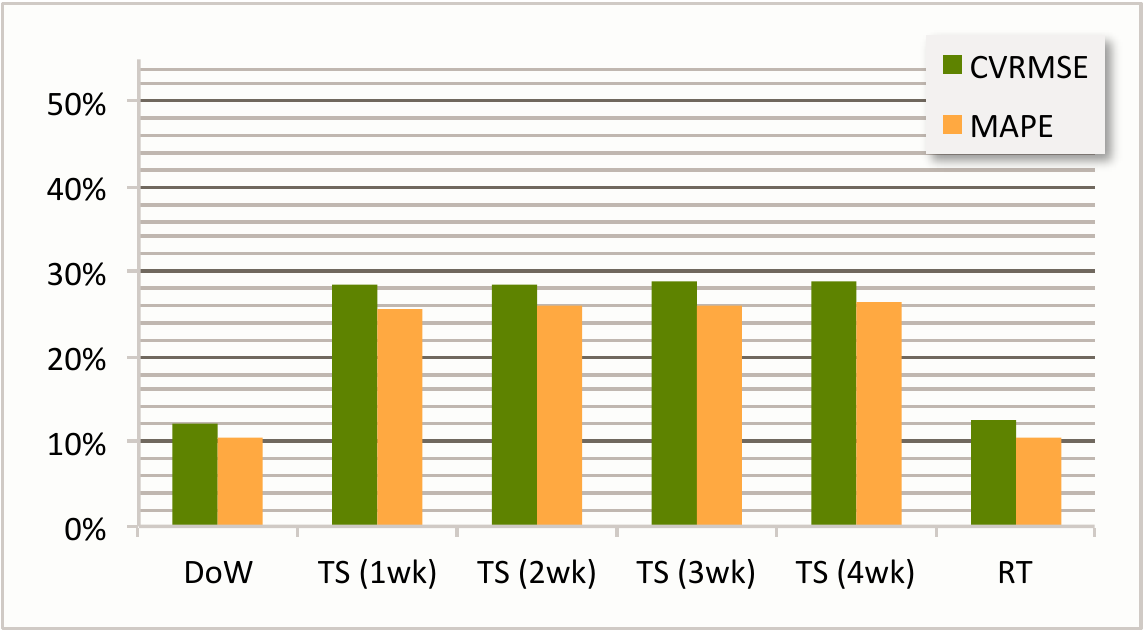}
   \caption{OFF}
	\label{fig:OHE24-CV-MAPE}
 \end{subfigure}
 \begin{subfigure}[b]{\textwidth}
   \includegraphics[width=\textwidth] {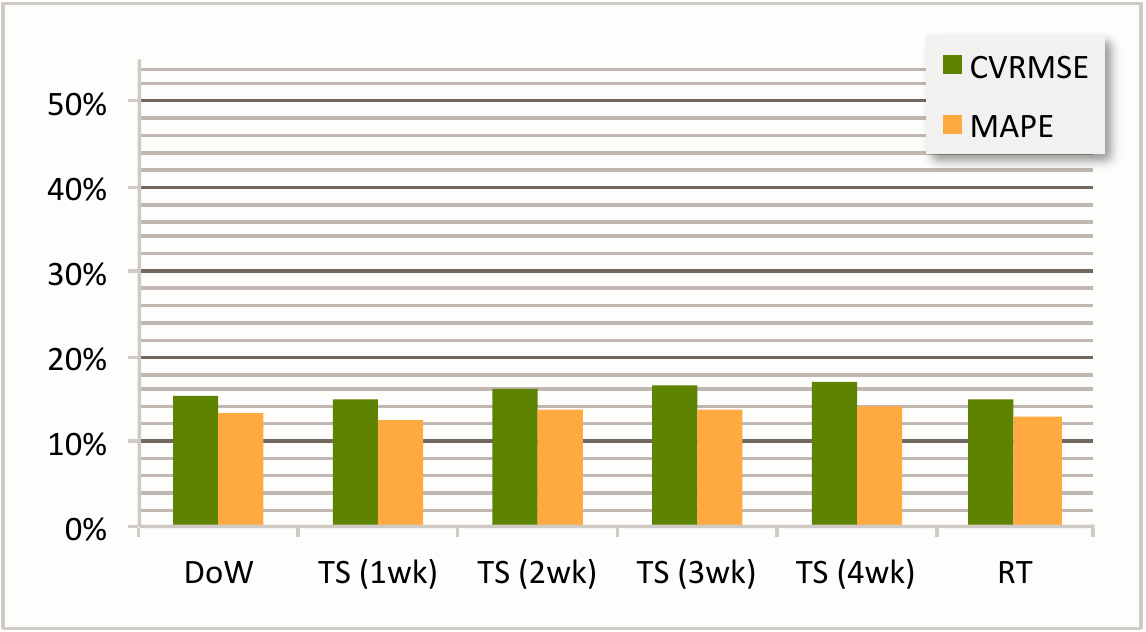}
   \caption{ACD}
   	\label{fig:KAP24-CV-MAPE}
 \end{subfigure}

\caption{\textbf{\emph{CVRMSE}} and \textbf{\emph{MAPE}} values for \textbf{\emph{24-hour predictions}} for campus and four buildings. Lower values
  are better. Day of Week (DoW) baseline, ARIMA Time Series (TS) with 1, 2, 3 \&
  4-week\del{ ahead} prediction horizons, and Regression Tree \add{(RT)} models are on
  X-axis. Campus \del{shows}\add{has} the
  smallest errors, \del{the}RES residential building the largest, and, except for OFF,\del{the} TS \del{or}\add{and} RT
  outperform the baseline.}
\end{minipage}
~
\begin{minipage}[b]{0.33\textwidth}
\centering
  \begin{subfigure}[b]{\textwidth}
\centering
   \includegraphics[width=\textwidth] {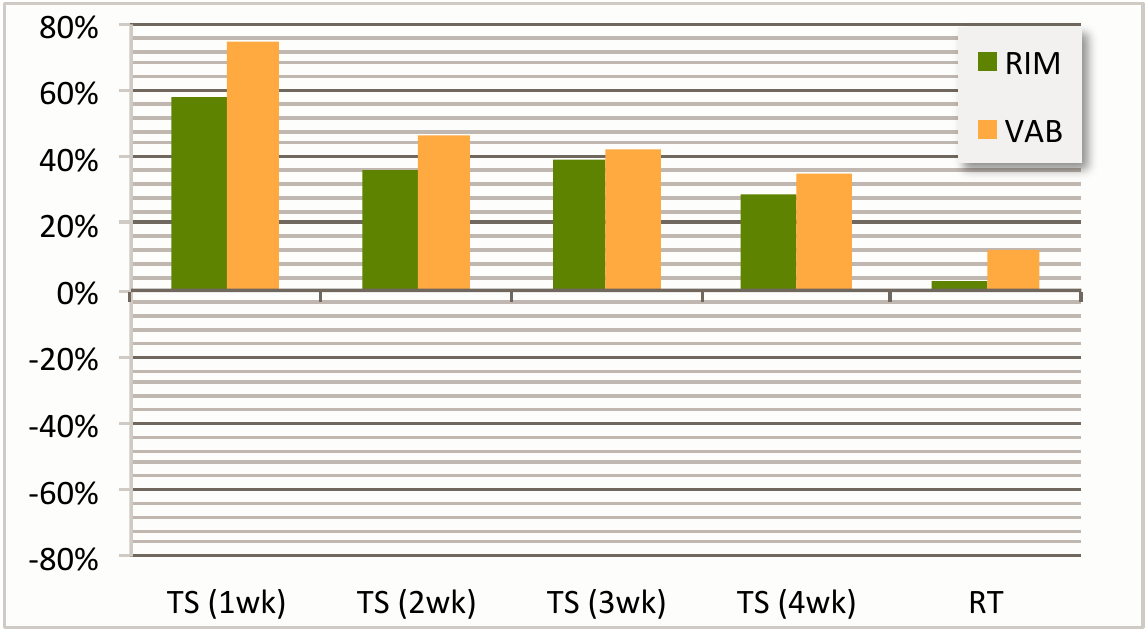}
   \caption{Campus}
   	\label{fig:C24-RIM-VAB}
 \end{subfigure}
 \begin{subfigure}[b]{\textwidth}
   \includegraphics[width=\textwidth] {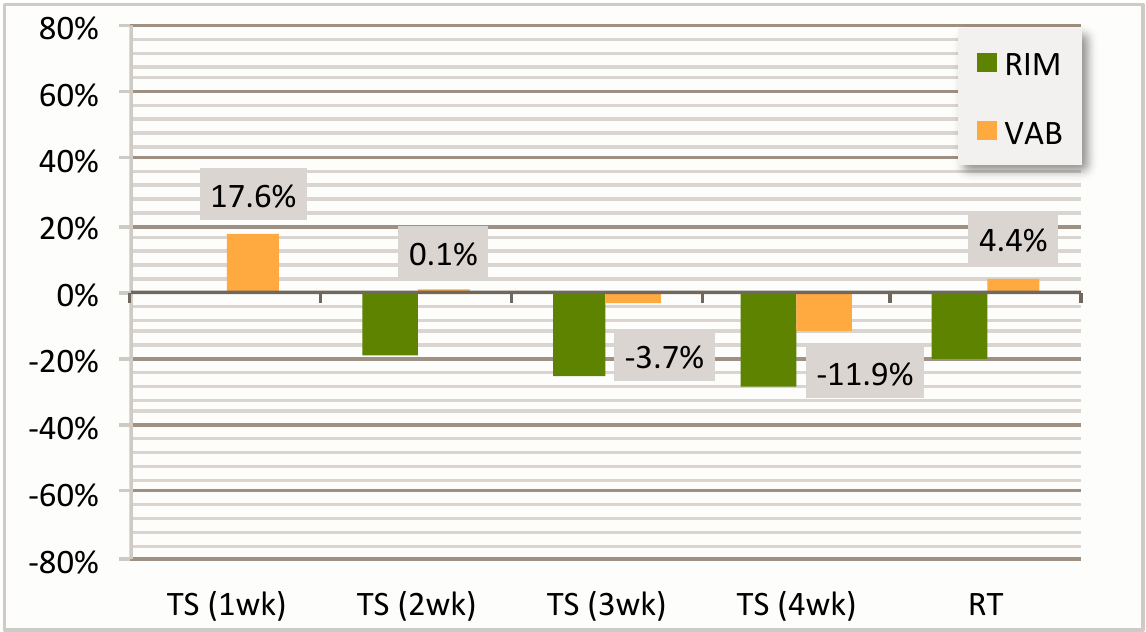}
   \caption{DPT}
      	\label{fig:MHP24-RIM-VAB}
 \end{subfigure}
 \begin{subfigure}[b]{\textwidth}
   \includegraphics[width=\textwidth] {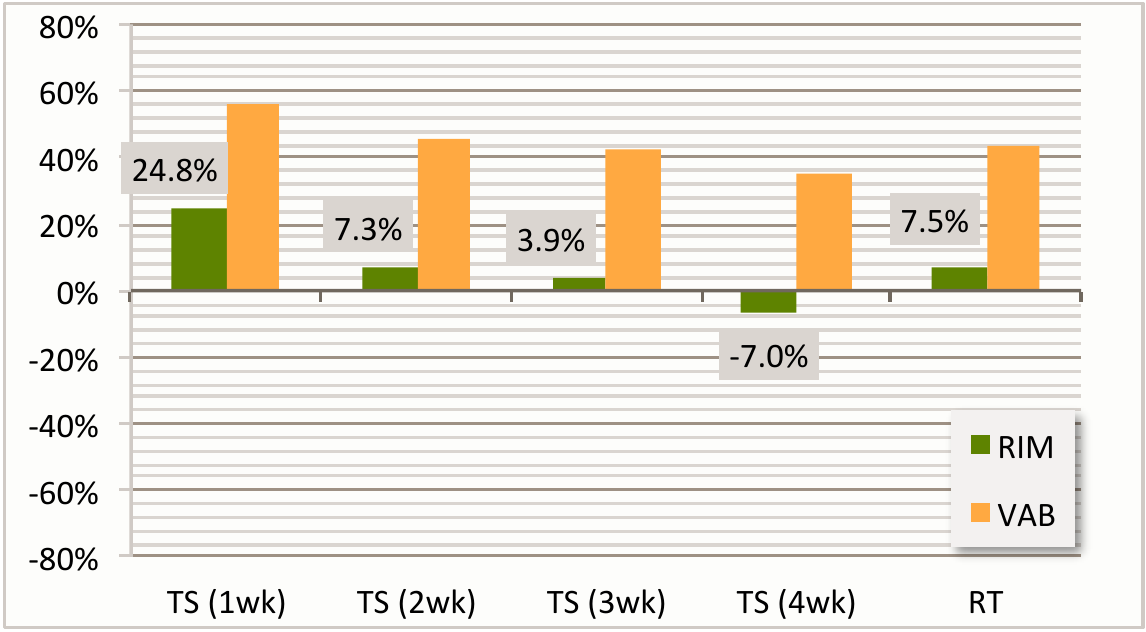}
   \caption{RES}
 \label{fig:FLT24-RIM-VAB}
 \end{subfigure}
 \begin{subfigure}[b]{\textwidth}
   \includegraphics[width=\textwidth] {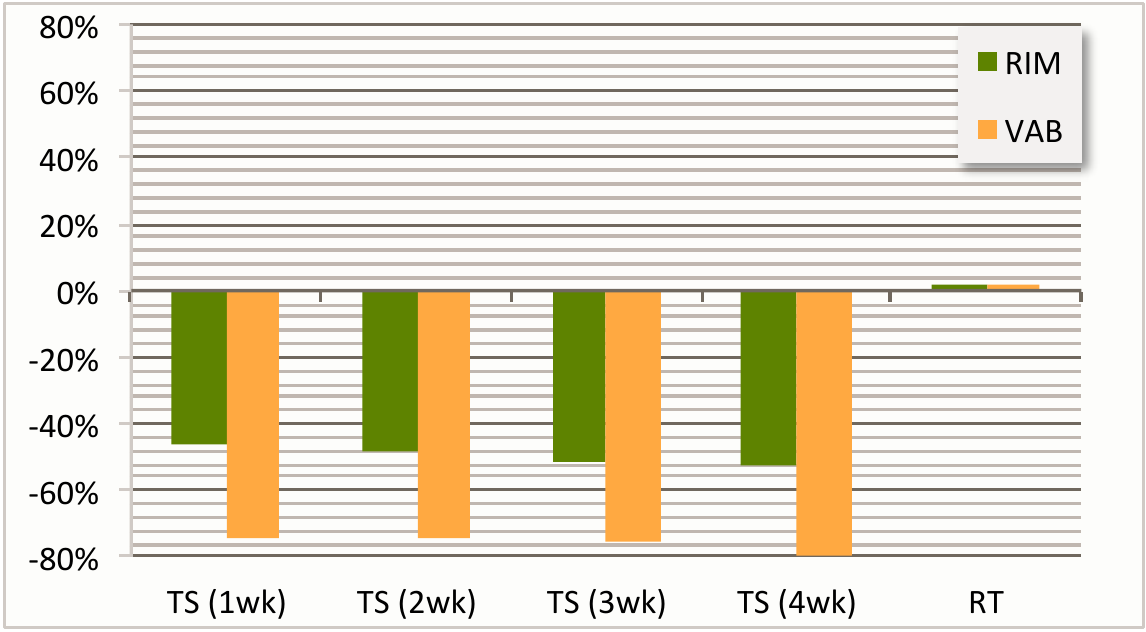}
   \caption{OFF}
 \label{fig:OHE24-RIM-VAB}
 \end{subfigure}
 \begin{subfigure}[b]{\textwidth}
   \includegraphics[width=\textwidth] {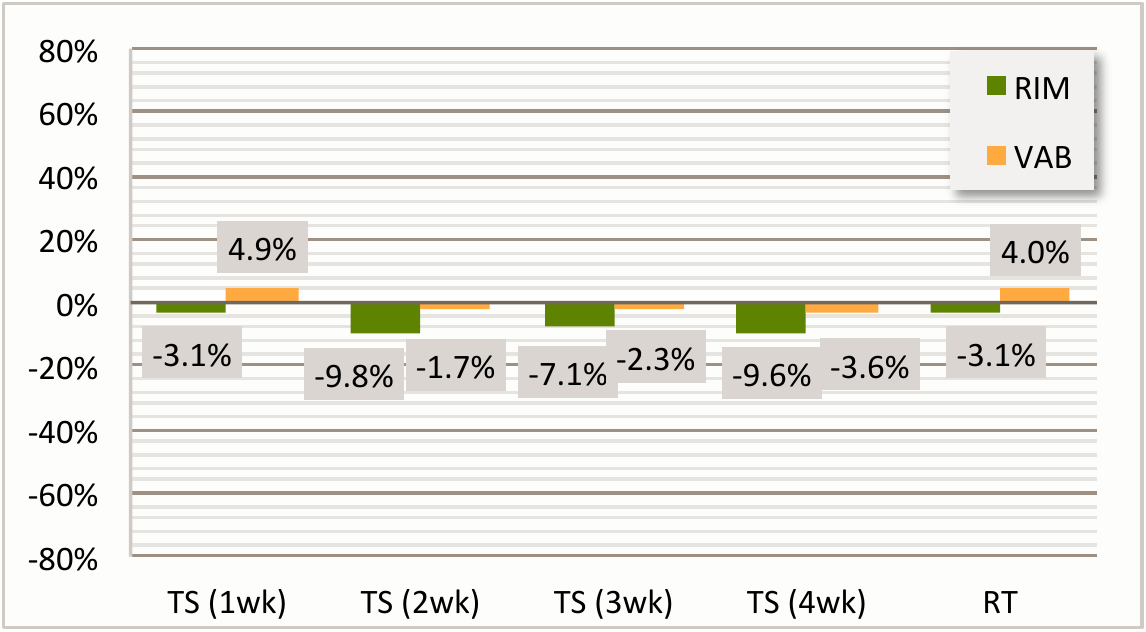}
   \caption{ACD}
 \label{fig:KAP24-RIM-VAB}
 \end{subfigure}

\caption{Relative Improvement (\textbf{\emph{RIM}}) and Volatility-Adjusted Benefit (\textbf{\emph{VAB}}) values for \textbf{\emph{24-hour
  predictions}} for campus and four buildings. Higher values indicate better performance
  relative to DoW baseline; zero value \del{indicates}\add{means} performance similar to baseline. DoW is more
  volatile for RES due to summer vacation.\del{, and the} VAB for RT and TS are high, \del{exhibiting}\add{showing} resilience.}

\end{minipage}
 ~
\begin{minipage}[b]{0.33\textwidth}
\centering
  \begin{subfigure}[b]{\textwidth}
\centering
   \includegraphics[width=\textwidth] {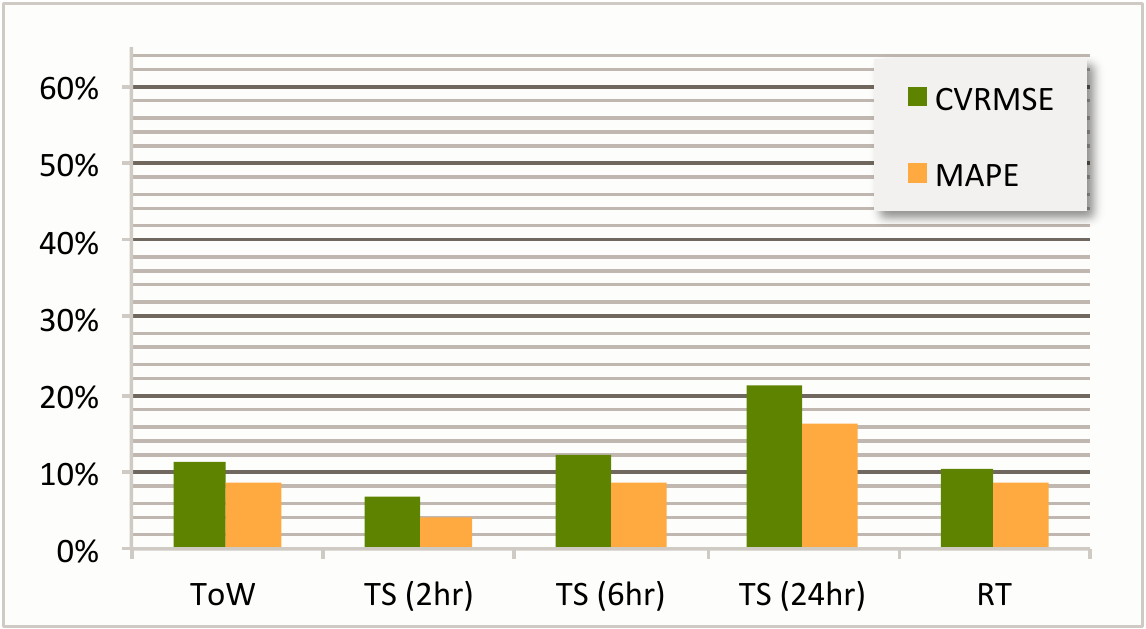}
   \caption{Campus}
\label{fig:C15-CV-MAPE}
 \end{subfigure}
 \begin{subfigure}[b]{\textwidth}
   \includegraphics[width=\textwidth] {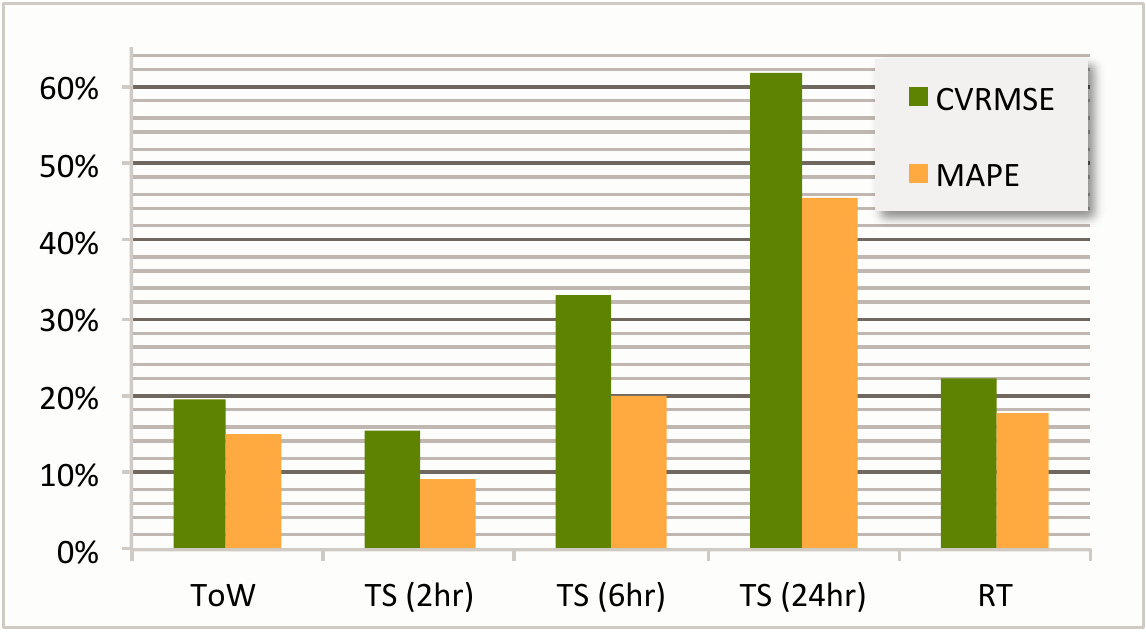}
   \caption{DPT}
 \label{fig:MHP15-CV-MAPE}
 \end{subfigure}
 \begin{subfigure}[b]{\textwidth}
   \includegraphics[width=\textwidth] {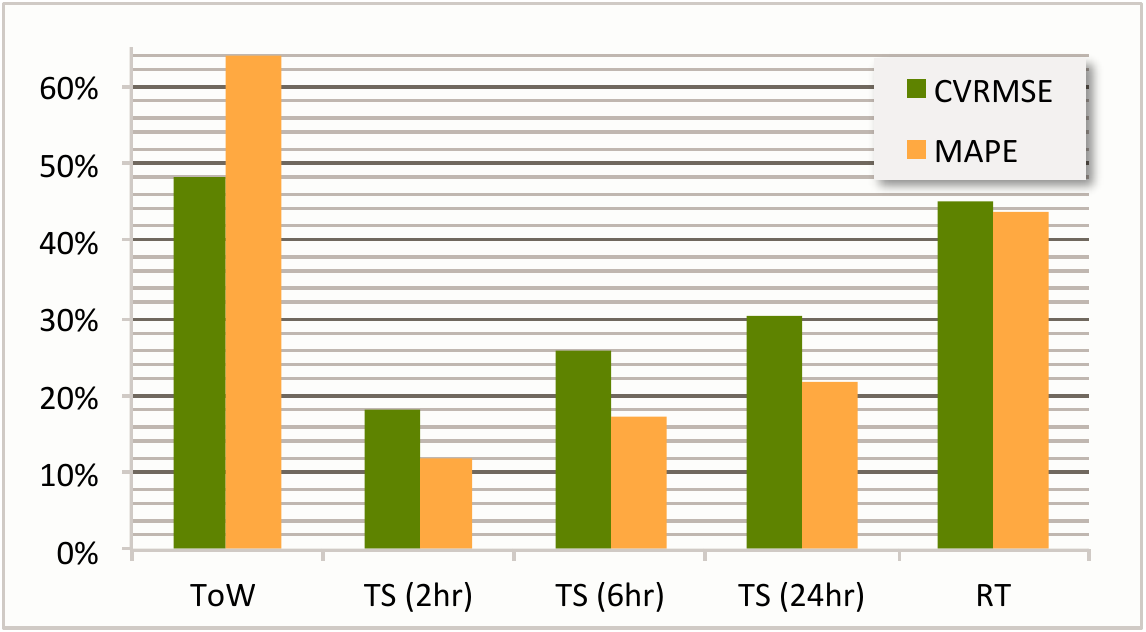}
   \caption{RES}
 \label{fig:FLT15-CV-MAPE}
 \end{subfigure}
 \begin{subfigure}[b]{\textwidth}
   \includegraphics[width=\textwidth] {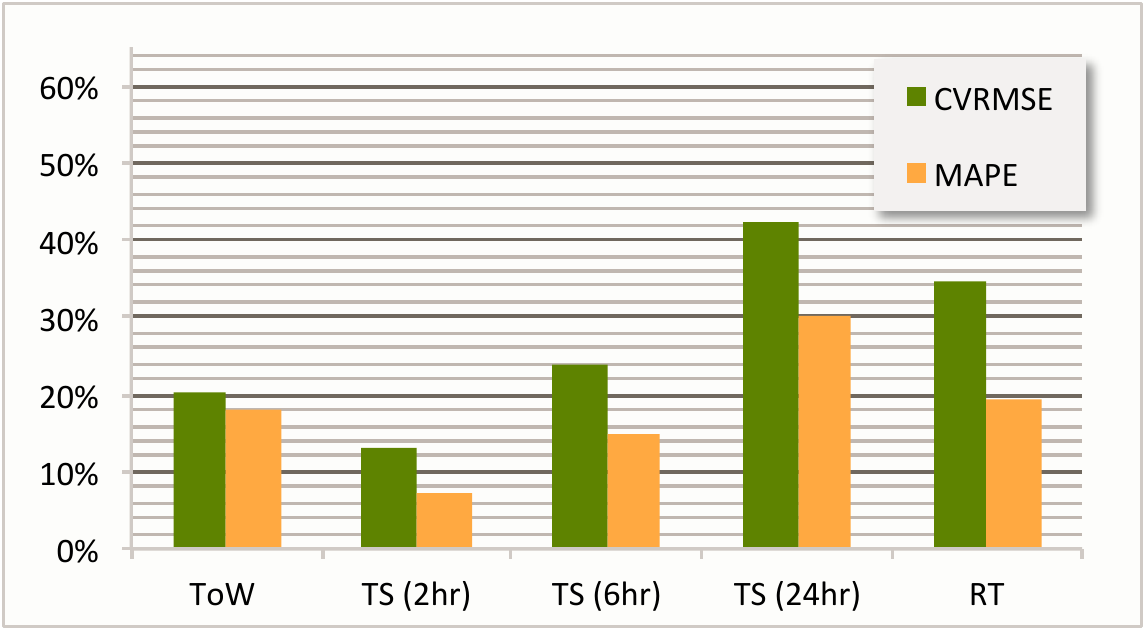}
   \caption{OFF}
 \label{fig:OHE15-CV-MAPE}
 \end{subfigure}
 \begin{subfigure}[b]{\textwidth}
   \includegraphics[width=\textwidth] {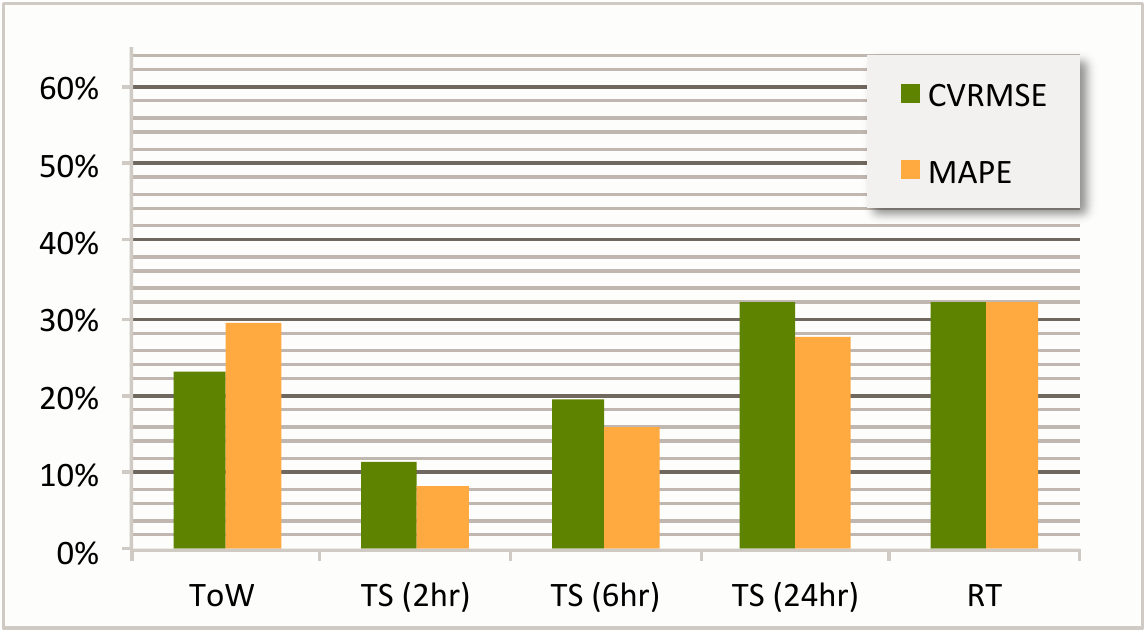}
   \caption{ACD}
 \label{fig:KAP15-CV-MAPE}
 \end{subfigure}

\caption{\textbf{\emph{CVRMSE}} and \textbf{\emph{MAPE}} values for \textbf{\emph{15-min
      predictions}} for campus and four buildings. Lower errors indicate better model performance. \del{ToW is the Time of Week baseline; TS (2hr)
    is the time series model that makes predictions for the next 2-hours, others are similarly
    defined; RT is the regression tree model.}\add{Time of Week baseline (ToW), ARIMA Time Series
    (TS) at 2, 6 \& 24-hour horizons, and Regression Tree (RT) models are shown.} Errors usually increase as the prediction horizon is increased. RT is independent of prediction horizon.}
\label{fig:CV-MAPE-15}
\end{minipage}

\end{figure*}

Fig.~\ref{fig:C24-CV-MAPE} presents the CVRMSE and MAPE measures for the DoW baseline, RT, and TS
models, the latter for four different horizons, for campus 24-hour predictions. By these measures,
TS models at different horizons offer higher accuracy than the RT and DoW models. This is understandable, given the noticeable difference in mean and standard deviations (Table~\ref{table:dataset}) between the training and test periods. TS \del{iteratively}\add{incrementally} uses \textit{more recent data as a moving window}, while RT and DoW model are only
trained on the two years' \add{test} data. Also, the errors for TS deteriorate as
the prediction horizon increases. This is a consequence of their dependence on recent lag values,
making them suited only for \textit{near-term predictions}. RT models are \textit{independent of
  prediction horizons} (assuming future feature values are \del{deterministic}\add{known}), and therefore preferable
for predictions with long horizons. The DoW\del{ baseline} errors are \emph{marginally} higher than
RT. This is quickly evident using our relative improvement (RIM) measure
(Fig.~\ref{fig:C24-RIM-VAB}), that reports an improvement of $2.5\%$ for RT and $58.39\%$ for TS
(1wk) over the baseline. However, when volatility is accounted for, this margin over the DoW
increases to a VAB of $11.45\%$ and $74.42\%$ for RT and TS (1wk), making them \emph{much more dependable}.


\vspace{-0.05in}
\subsection{24-hour Building Predictions}
The CVRMSE and MAPE measures for DPT (Fig. \ref{fig:MHP24-CV-MAPE}) diverge in their ranking of
the RT and TS models; RT is best based on CVRMSE while TS (1wk) is best on MAPE. This
divergence highlights the value of having different error measures. In CVRMSE, residual errors
are squared and thus \textit{large errors are magnified} more than in MAPE. Our RIM measure offers
another perspective as a relative measure independent of error values
(Fig.~\ref{fig:MHP24-RIM-VAB}). TS (1wk) \emph{is clearly more favorable} than RT, performing better
than the baseline in $50\%$ of predictions (RIM $\approx 0$) compared to RT (RIM=$-19.88\%$). When accounting for
volatility in VAB, TS (1wk) outperforms the DoW (VAB=$17.62\%$) and even RT exhibits lesser
relative volatility (VAB=$4.35\%$). These demonstrate why multiple measures offer a more holistic view
of the model performance.

RES has 100's of residential suites with independent power controls, and hence
higher consumption variability. This accounts for the higher errors in predictions across
models (Fig.~\ref{fig:FLT24-CV-MAPE}). Further, the building is \emph{unoccupied during
summer} and vacation periods. Hence, it is unsurprising to see DoW perform particularly
worse. (We verified the impact of summer by comparing DoW with DoY. DoY did
perform better, but for consistency, we retain DoW as the baseline.)  RT has lower errors than
DoW as it captures \textit{schedule-related features like holidays and summer semester}. However,
the test data for RES has a smaller mean than the training data (Table~\ref{table:dataset}), thus
skewing predictions. TS (1wk) has the smallest error due to its ability to capture \textit{changing
and recent trends}.  The RIM (Fig.~\ref{fig:FLT24-RIM-VAB}) with respect to DoW is greater than 0
for all models except TS (4wk), which performs worse than the baseline $7\%$ of the time, even as it has
comparatively smaller errors. Given the high \emph{consumption variability} for RES, performance
under volatility
is important. A high VAB is desirable and provided by all models.

For OFF (Fig.~\ref{fig:OHE24-CV-MAPE}), we again see a divergence in model ranking when based on
CVRMSE or on MAPE, reflecting the benefit of each measure. Uniquely, neither RT nor TS are able to
surpass the DoW baseline in terms of CVRMSE. We independently verified if the consumption pattern of
this building is highly-correlated with the DoW by examining the decision tree generated by RT, the
best choice in terms of MAPE. We found the DoW feature to be present in the \emph{root node of the
  tree} while the holiday flag was at the second level. RT is also the only model which
(marginally) outperforms the baseline on RIM and VAB (Fig.~\ref{fig:OHE24-RIM-VAB}), thus delivering
the benefits of using a feature-based approach that subsumes DoW. TS fails to do well, possibly
due to \textit{temporal dependencies} that extend beyond the 7-day lag period\del{ chosen}. 
\add{It is notable that while DoW is the preferred model based on CVRMSE
  (Fig.~\ref{fig:OHE24-CV-MAPE}) for OFF, measures we propose, such as RIM and VAB (Fig.~\ref{fig:OHE24-RIM-VAB}) that evaluate performance against the DoW baseline, indicate that RT is the better choice.}

For ACD (Fig.~\ref{fig:KAP24-CV-MAPE}), TS (1wk) and RT perform incrementally better than DoW on
CVRMSE and MAPE, and VAB is positive for only these two models (Fig.~\ref{fig:KAP24-RIM-VAB}). The
sharp change in standard deviation between the training and test data accounts for the
higher sensitivity to volatility of the baseline (Table~\ref{table:dataset}). \del{However,}\add{But} we observe
slightly negative values of RIM for all models, implying \emph{more frequent errors than the
baseline}.
 
\subsection{15-min Campus Predictions}

\begin{figure*}
        \centering
        \begin{subfigure}[b]{0.3\textwidth}
                \centering
                \includegraphics[width=\textwidth]{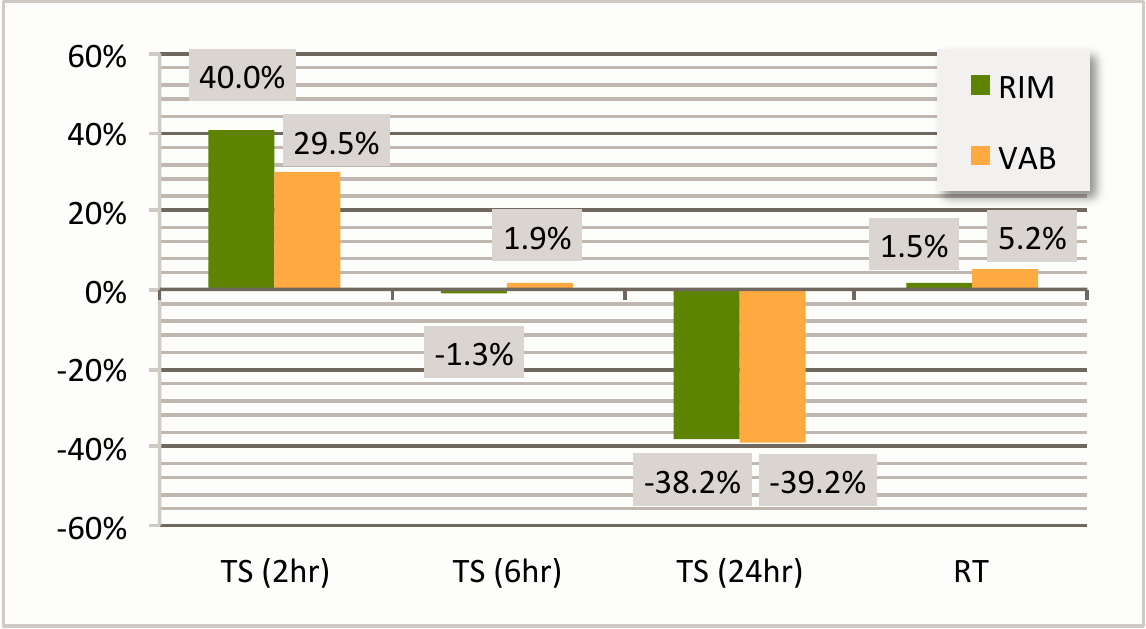}
                \caption{Campus}
                \label{fig:C15-RIM-VAB}
        \end{subfigure}%
        ~ 
        \begin{subfigure}[b]{0.3\textwidth}
                \centering
                \includegraphics[width=\textwidth]{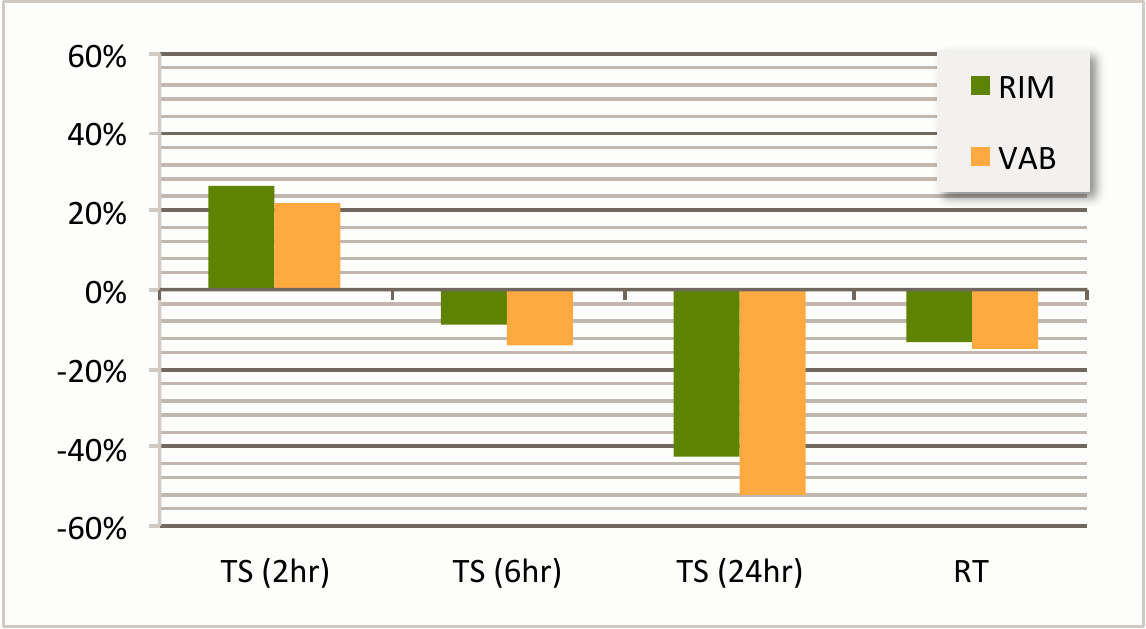}
                \caption{DPT}
                \label{fig:MHP15-RIM-VAB}
        \end{subfigure}
        ~ 
        \begin{subfigure}[b]{0.3\textwidth}
                \centering
                \includegraphics[width=\textwidth]{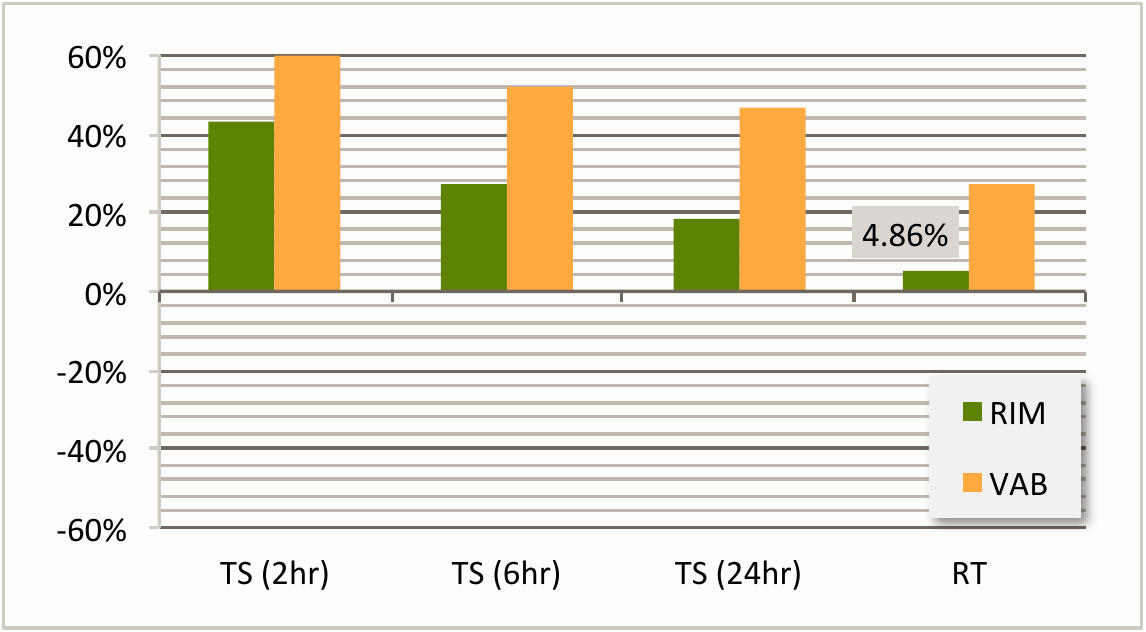}
                \caption{RES}
                \label{fig:FLT15-RIM-VAB}
        \end{subfigure}
        \begin{subfigure}[b]{0.3\textwidth}
                \centering
                \includegraphics[width=\textwidth]{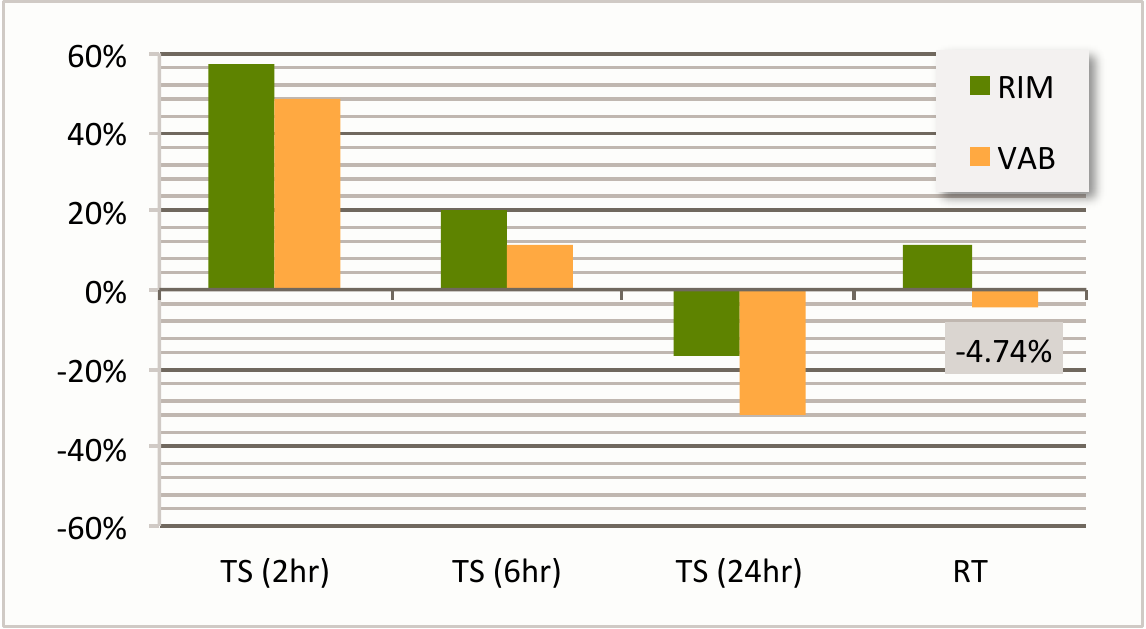}
                \caption{OFF}
                \label{fig:OHE15-RIM-VAB}
        \end{subfigure}
        \begin{subfigure}[b]{0.3\textwidth}
                \centering
                \includegraphics[width=\textwidth]{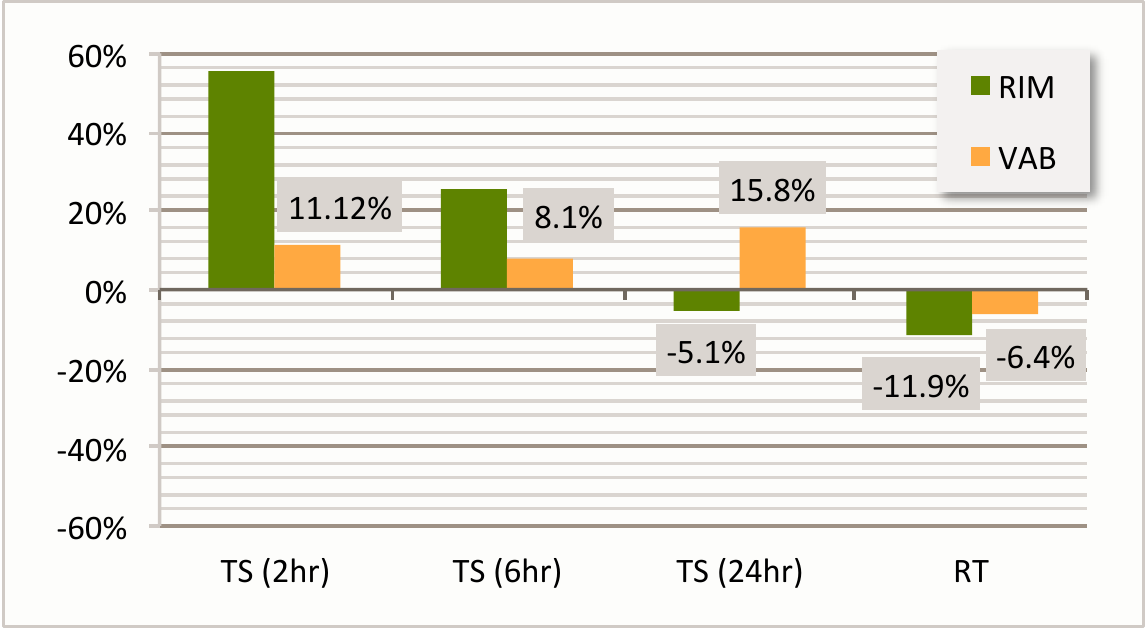}
                \caption{ACD}
                \label{fig:KAP15-RIM-VAB}
        \end{subfigure}
        \caption{\textbf{\emph{RIM}} and \textbf{\emph{VAB}} values for \textbf{\emph{15-min predictions}}
          for campus and four buildings. Higher values indicate better model performance with respect to the
          baseline; zero indicates similar performance as baseline. TS (2hr) usually offers highest
          reliability in all cases.}
        \label{fig:RIM-VAB-15}
\end{figure*}

The 15-min predictions for the \del{Campus}\add{campus} shows TS (2hr) to fall closest to the observed values, based on
CVRMSE ($6.88\%$) and MAPE ($4.18\%$) (Fig.~\ref{fig:C15-CV-MAPE}). This accuracy is validated
relative to the baseline, with high RIM and VAB values 
(Fig. \ref{fig:C15-RIM-VAB}). These reflect
the twin benefits of \emph{large spatial granularity} of the campus, which make its consumption slower
changing, and the short horizon of TS (2hr), helping it capture \emph{temporal similarity}. RT is
the next best, performing similar to TS (6hr) and ToW baseline on CVRMSE, MAPE and RIM, though it is
better with volatility (VAB=$5.21\%$). 

\subsection{15-min Building Predictions}
For 15-min predictions for buildings, we see that TS (2hr) is the only candidate model that
always does better than the ToW baseline on all four measures (Figs.~\ref{fig:MHP15-CV-MAPE}--\ref{fig:KAP15-CV-MAPE} \&
\ref{fig:MHP15-RIM-VAB}--\ref{fig:KAP15-RIM-VAB}). TS (6hr) and RT are occasionally better than ToW on CVRMSE and MAPE, and TS (24hr)
rarely. Their CVRMSE errors are also uniformly larger than MAPE, showing that the models suffer more from
occasional large errors. The academic environment with weekly class schedules encourages a uniform
energy use behavior based on ToW, that is hard to beat. RES is the\del{ only} exception, where all
candidate models are better than the baseline (Fig.~\ref{fig:FLT15-CV-MAPE}), given the aberrant
summer months when \del{the building}\add{it} is unused. 

However, when we consider the RIM and VAB measures, it is interesting to note that the candidate
models are not \del{that bad relative to}\add{significantly worse than} the baseline (Fig.~\ref{fig:MHP15-RIM-VAB}--\ref{fig:KAP15-RIM-VAB}). In fact, TS
(6hr) is better than ToW for all buildings but DPT, showing that it is more often accurate and more
reliable under volatility. RT, however, is more susceptible to volatility and shows negative values
for all buildings but RES. While TS (2hr) followed by TS (6hr) are obvious choices for short horizons\del{predictions}, ToW and RT have the advantages of being able to predict over a longer term. In the
latter case, ToW actually turns out to be a better model.

\subsection{Cost Measures}

The data and compute cost measures, discussed here, are orthogonal to the other
application-independent measures, and their values are summarized in Table~\ref{table:indep_cost}. \add{Making cost assessment helps grid managers ensure rational use of resources, including skilled manpower and compute resources, in the prediction process.}

\begin{table}[!t]
\renewcommand{\arraystretch}{1.0}

\caption{\textbf{Application-independent Cost Measures}. Prediction horizon is 4~weeks for 24-hour
  predictions; and 24~hours for 15-min predictions. CD measures number of unique feature values used
  in training and testing. TS and baseline \del{don't}\add{do not} have a training cost.}
\centering  
\begin{tabular}{l r r r} 

\hline\hline   
& Data Cost & \multicolumn{2}{c}{Compute Cost \emph{(millisec)}}\\               
Model & $CD$ & $CC_{t}$  & $CC_{p}$ \\ 
\hline
\textbf{DoW/ToW Baseline} & & &\\
24-hour predictions & 1,096 & - & - \\
15-min predictions & 1,05,216 & - & - \\
\hline
\textbf{Time Series} & & &\\
24-hour predictions & 1,096 & - & 101 \\
15-min predictions & 1,05,216 & - & 933 \\
\hline
\textbf{Regression Tree} & & &\\
24-hour predictions & 3,301  & 94 & 1.6 \\
15-min predictions &  1,31,629 & 17,275 & 48 \\
\hline
\end{tabular}
\label{table:indep_cost} 
\vspace{-0.2in}
\end{table}

\textbf{Data Cost (CD):}
The baselines and TS are univariate models that require only the electricity
consumption values for the training and test periods. Hence their data costs are smaller, and
correspond to the number of intervals trained and tested over. RT model has a higher cost due to the
addition of several features (\S~\ref{config}). However, the cost does not increase linearly with
the number of features and instead depends on the number of unique feature values. As a result, its
\del{CD}\add{data cost} is only $\sim 25\%$ and $\sim 300\%$ greater than TS for 24-hr and 15-min
predictions \add{respectively}.

\textbf{Compute Cost (CC):}
We train over 2 years and predict for 4~weeks (24-hour granularity) and 24~hours (15-min) on a 
Windows Server with AMD~3.0GHz CPU and 64GB RAM, and report the average over 10 experiment runs. The baseline's compute cost is trivial
as it is just an average over past values, and we ignore it. For the TS models, \del{training}\add{retraining} is
interleaved with the prediction and we report them 
as part of the prediction cost ($CC_p$). We found prediction times for TS to be identical
across campus and the four buildings, and the 15-min predictions to be $\sim9\times$ the cost of
24-hour -- understandable since there are $\sim10\times$ the data points.  The horizons did not
affect these times. For RT, we find the
training and prediction times to be similar (but not same) across campus and four buildings, and
this is seen in the differences in the sizes of the trees constructed. We
report their average time. While RT has a noticeable training time (17~secs for 15-min), its
prediction time is an order of magnitude smaller than TS. As a result, its regular use for
prediction is cheaper\del{, and it}\add{. It} is more responsive, with a lower prediction latency, even as the number of
buildings (or customers) increase to the thousands.

\section{Analysis of Dependent Measures}
\label{Sec:dep_evaluation}

\del{While the earlier measures offered a broad evaluation of the models based on multiple dimensions,
our four application dependent measures (\S~\ref{Sec:app_spec_measures}) provide deeper insight in the appropriateness of the models for
specific applications. The measures' parameter values for the three
smart grid applications (\S~\ref{Sec:use_cases}) are listed in Tables~\ref{table:parameters} \& \ref{table:cost}. Here,
we discuss the rationale for picking these values - guided by power systems and
sociology experts - and their use in model evaluation.}
\add{The application-dependent measures (\S~\ref{Sec:app_spec_measures}) enable model selection for
  specific \del{applications}\add{application scenarios}. For each application (\S~\ref{Sec:use_cases}), the measures' parameter values are defined in consultation with the domain experts. These values are listed in Tables~\ref{table:parameters} \& \ref{table:cost}.} 
\begin{table}[!t]
\renewcommand{\arraystretch}{1.0}

\caption{\textbf{Application Specific Parameters.} $\alpha, \beta$ are over- and under-prediction
  penalties for DBPE, and $e_{t}$ is the error tolerance for REL.} 
\centering
\begin{tabular}{l c c} 

\hline\hline                  
Application \& Prediction Type & DBPE ($\alpha, \beta$) & REL ($e_{t}$) \\ 
\hline
\textbf{Planning} & &\\
24-hour Buildings & 0.50, 1.50 & 0.15\\
24-hour Campus & 1.00, 1.00 & 0.10 \\
\hline
\textbf{Customer Education} &  &\\
24-hour Building & 0.75, 1.25 & 0.15 \\
15-min Buildings (6AM-10PM) & 1.50, 0.50 & 0.10 \\
\hline
\textbf{Demand Response} &  &\\
15-min Campus (1PM-5PM) & 0.50, 1.50 & 0.05 \\
15-min Buildings (1PM-5PM) & 0.50, 1.50 & 0.10 \\
\hline
\end{tabular}
\label{table:parameters} 
\end{table}
\subsection{Planning}
\emph{Planning} requires medium- and long-term consumption predictions at 24-hour granularities for the
campus and buildings, six times a year. The short horizon of TS (4 weeks) precludes its
use. So we only consider DoW and RT models, but do report TS results. 

\textbf{Campus:}
For campus-scale decisions, both over- and under- predictions can be punitive. The former will
lead to over-provisioning of capacity with high costs while the latter can cause
reduced usability of capital investments. Hence, for DBPE, we equally weight $\alpha = 1$ and $\beta = 1$,
whereby DBPE reduces to MAPE. We set $e_{t}=10\%$, a relatively lower tolerance, since even a small
swing in error~\% for a large consumer like USC translates to large shifts in kWh. 

\begin{figure*}[!t]
\vspace{0.3in}
\begin{minipage}[b]{0.33\textwidth}
\centering
  \begin{subfigure}[b]{\textwidth}
\centering
   \includegraphics[width=\textwidth] {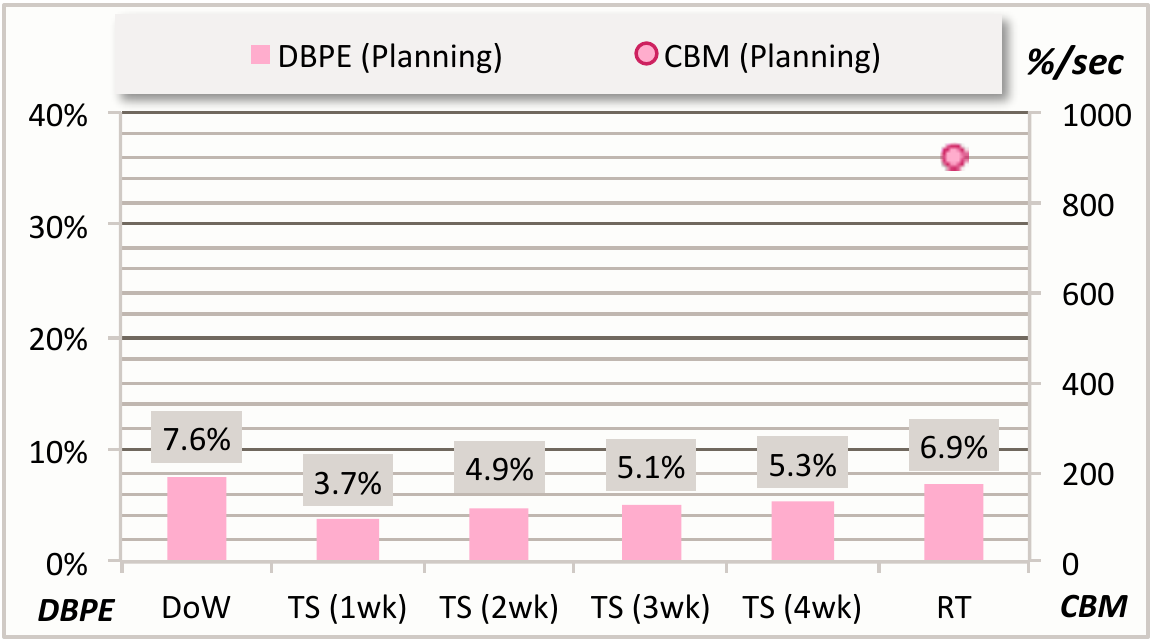}
   \caption{Campus}
   \label{fig:C24-DBPE}
 \end{subfigure}
 \begin{subfigure}[b]{\textwidth}
   \includegraphics[width=\textwidth] {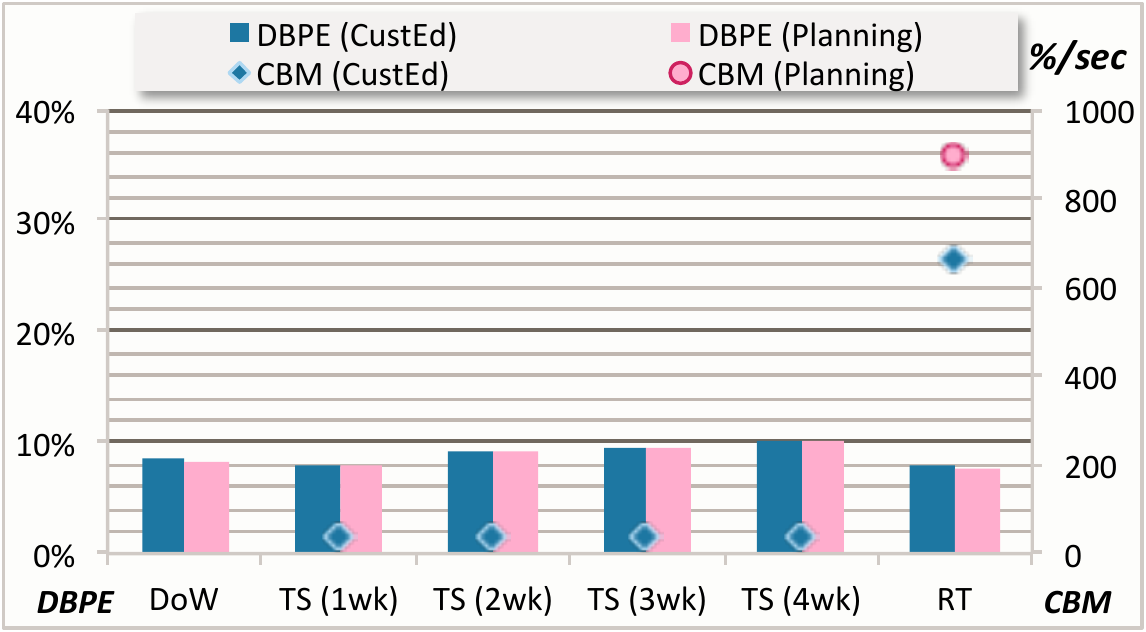}
   \caption{DPT}
   \label{fig:MHP24-DBPE}
 \end{subfigure}
 \begin{subfigure}[b]{\textwidth}
   \includegraphics[width=\textwidth] {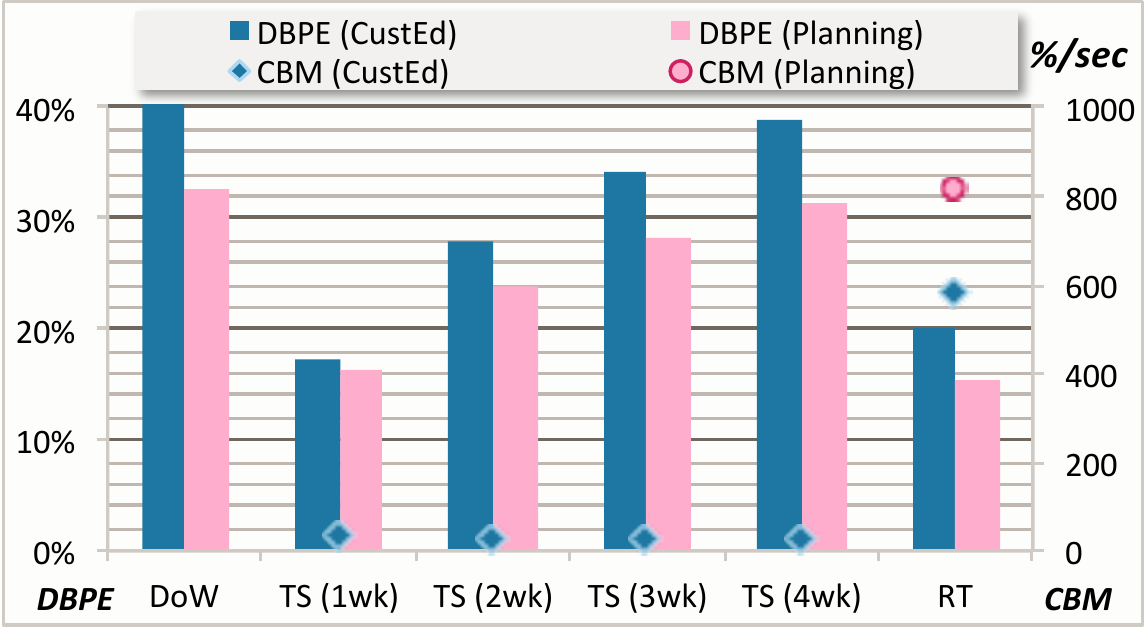}
   \caption{RES}
   \label{fig:FLT24-DBPE}
 \end{subfigure}
 \begin{subfigure}[b]{\textwidth}
   \includegraphics[width=\textwidth] {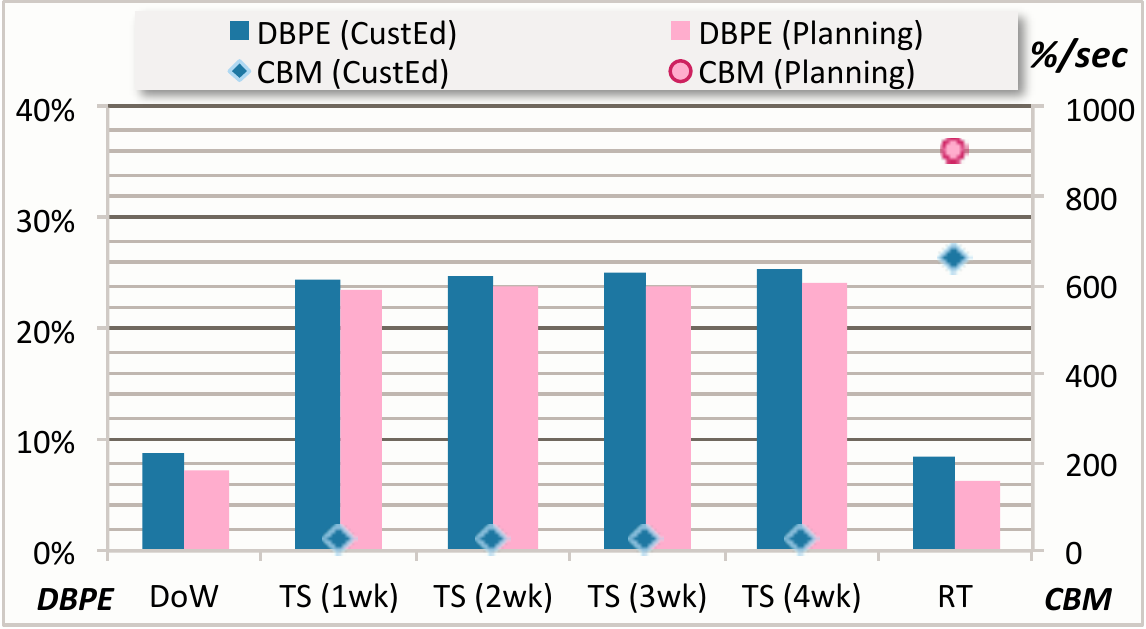}
   \caption{OFF}
   \label{fig:OHE24-DBPE}
 \end{subfigure}
 \begin{subfigure}[b]{\textwidth}
   \includegraphics[width=\textwidth] {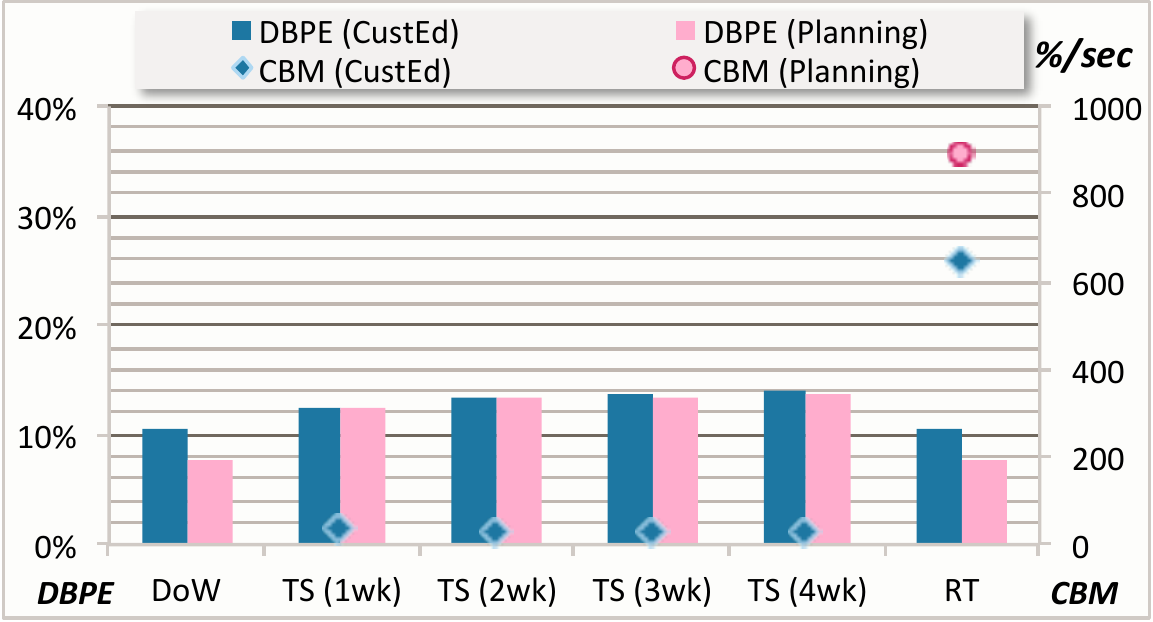}
   \caption{ACD}
   \label{fig:KAP24-DBPE}
 \end{subfigure}
\label{fig:DBPE24}
\caption{Domain bias percentage error (\textbf{\emph{DBPE}}), primary Y-axis, and Cost-Benefit
  Measure (\textbf{\emph{CBM}}), secondary Y-axis, for \textbf{\emph{24-hour predictions}} for
  Planning and Customer Education. Customer Ed. is not relevant for campus. Lower DBPE \&
  higher CBM are better, as seen in RT.} 
\end{minipage}
~
\begin{minipage}[b]{0.33\textwidth}
\centering
  \begin{subfigure}[b]{\textwidth}
\centering
   \includegraphics[width=\textwidth] {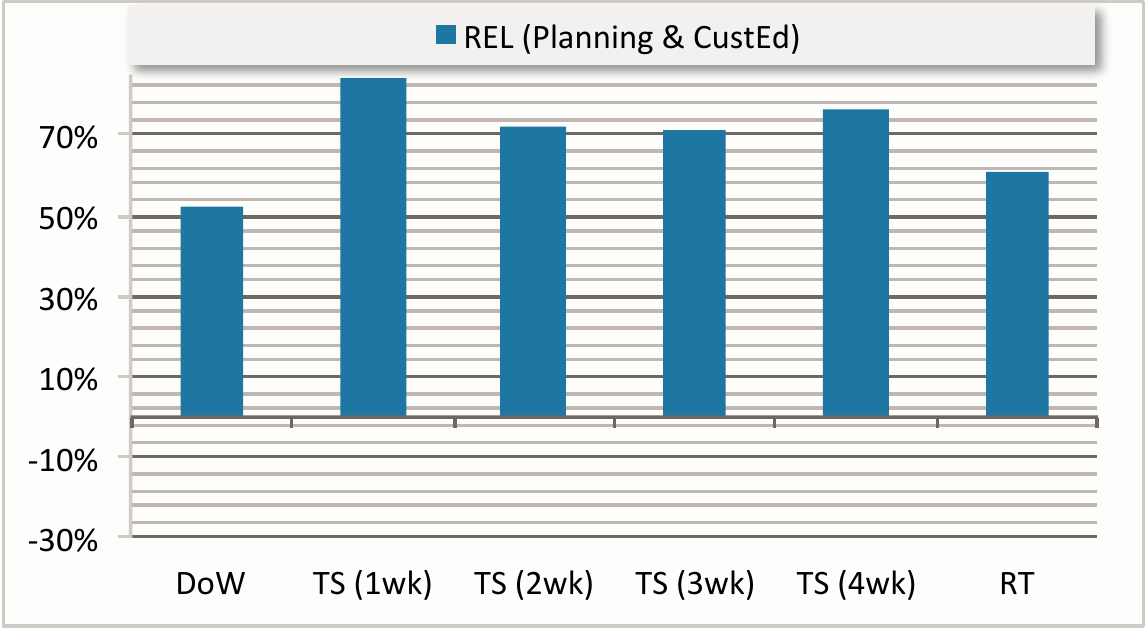}
   \caption{Campus}
   \label{fig:C24-REL}
 \end{subfigure}
 \begin{subfigure}[b]{\textwidth}
   \includegraphics[width=\textwidth] {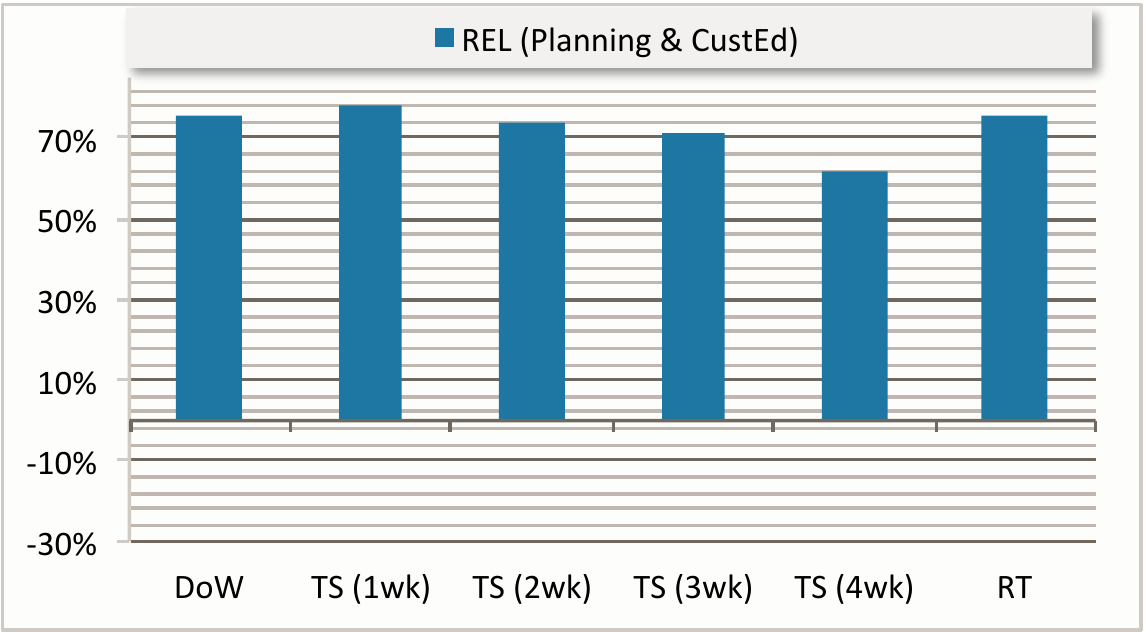}
   \caption{DPT}
    \label{fig:MHP24-REL}  
 \end{subfigure}
 \begin{subfigure}[b]{\textwidth}
   \includegraphics[width=\textwidth] {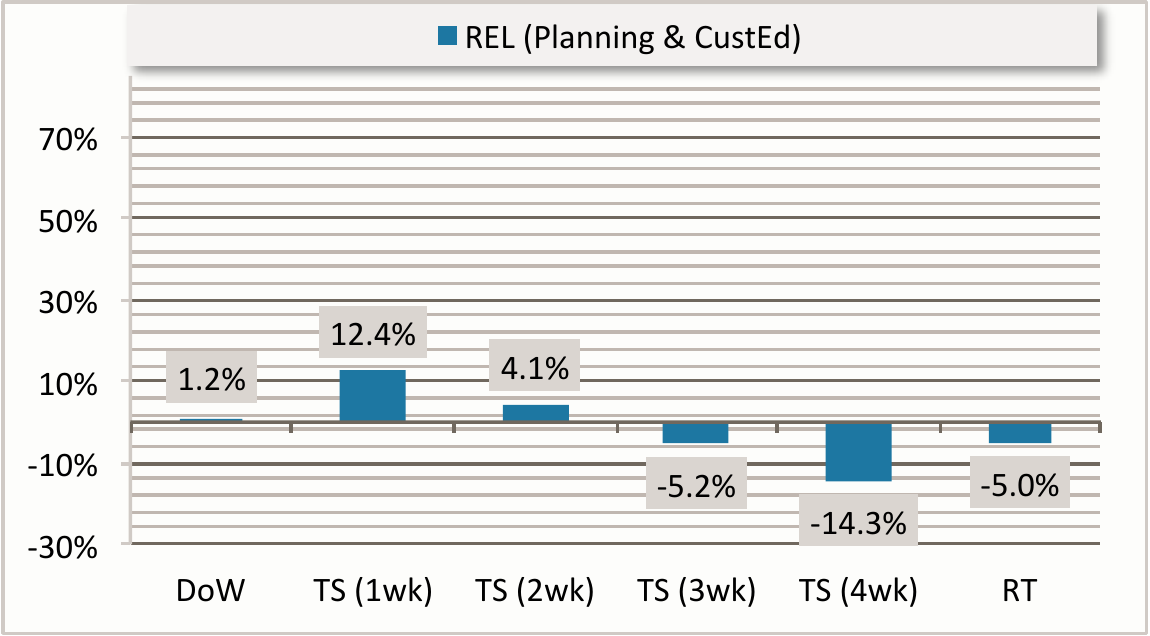}
   \caption{RES}
      \label{fig:FLT24-REL}
 \end{subfigure}
 \begin{subfigure}[b]{\textwidth}
   \includegraphics[width=\textwidth] {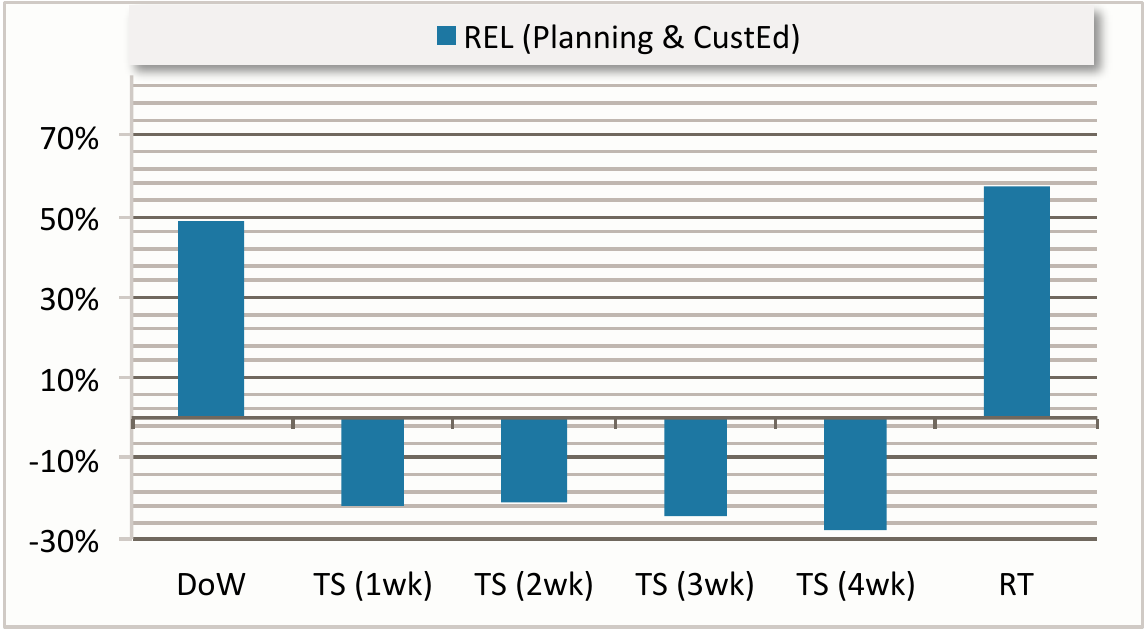}
   \caption{OFF}
      \label{fig:OHE24-REL}
 \end{subfigure}
 \begin{subfigure}[b]{\textwidth}
   \includegraphics[width=\textwidth] {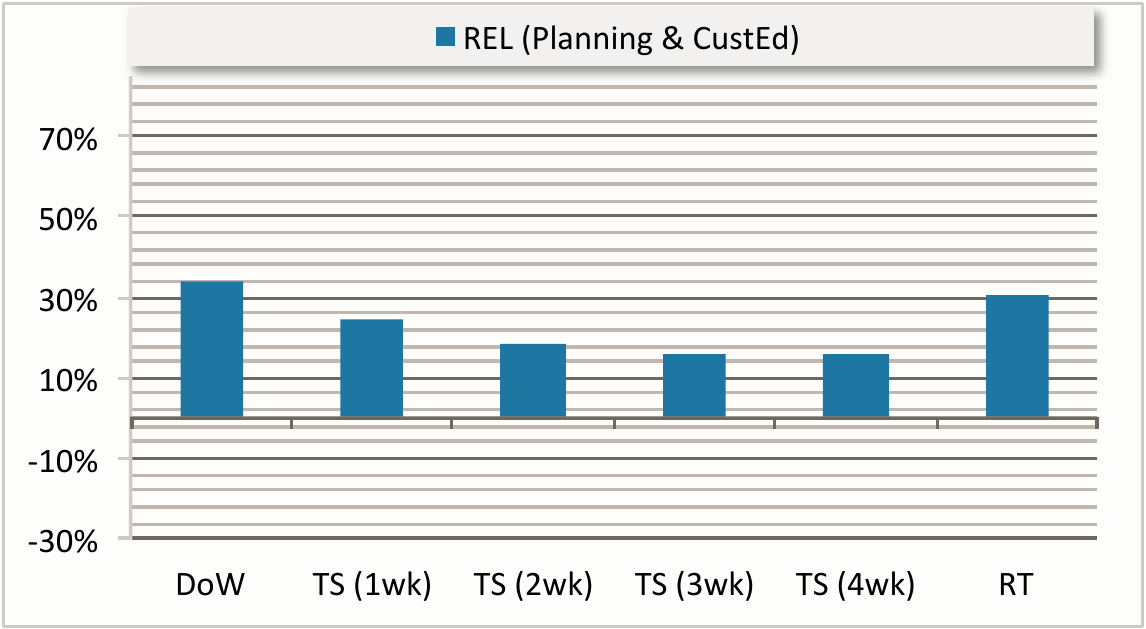}
   \caption{ACD}
      \label{fig:KAP24-REL}
 \end{subfigure}
\label{fig:REL24}
\caption{Reliability (\textbf{\emph{REL}}) values for \textbf{\emph{24-hour predictions}} for
  Planning, and Customer Education. Both have the same value of error tolerance parameter, and shown by
  a single graph. Higher values indicate better performance than the baseline; zero matches the baseline.}
\end{minipage}
~
\begin{minipage}[b]{0.33\textwidth}
\centering
  \begin{subfigure}[b]{\textwidth}
\centering
   \includegraphics[width=\textwidth] {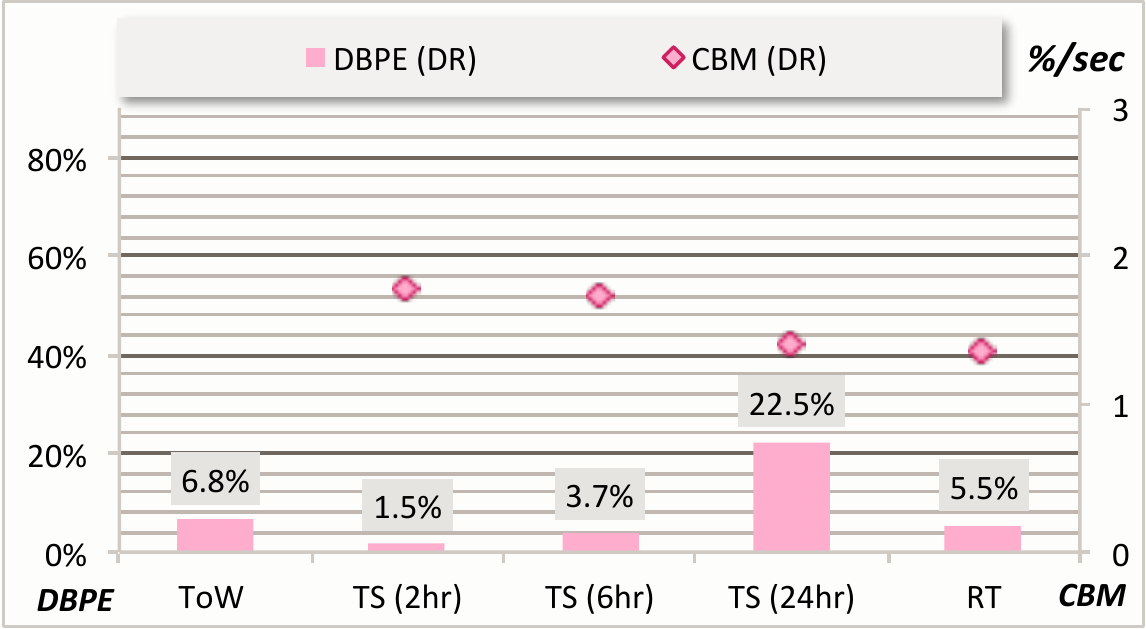}
   \caption{Campus}
      \label{fig:C15-DBPE}
 \end{subfigure}
 \begin{subfigure}[b]{\textwidth}
   \includegraphics[width=\textwidth] {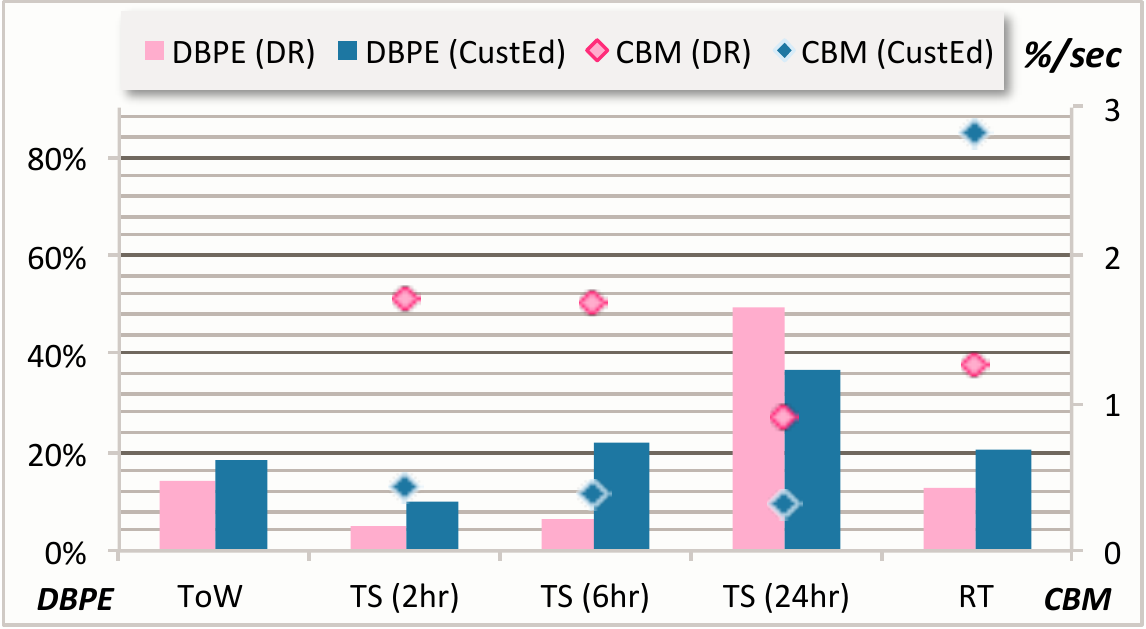}
   \caption{DPT}
      \label{fig:MHP15-DBPE}   
 \end{subfigure}
 \begin{subfigure}[b]{\textwidth}
   \includegraphics[width=\textwidth] {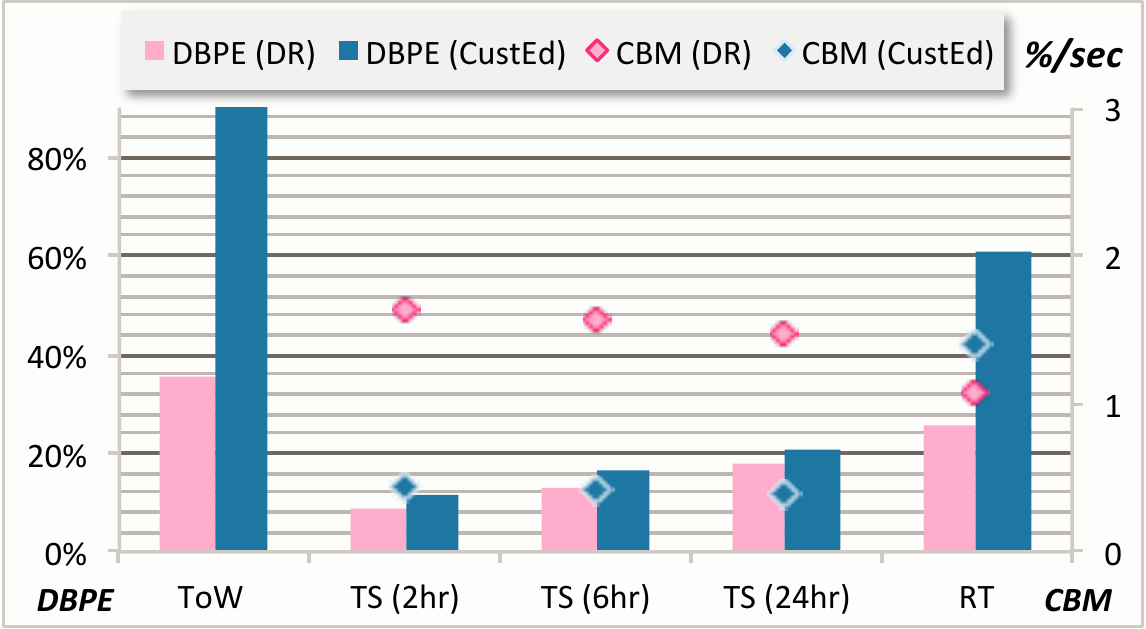}
   \caption{RES}
     \label{fig:FLT15-DBPE}
 \end{subfigure}
 \begin{subfigure}[b]{\textwidth}
   \includegraphics[width=\textwidth] {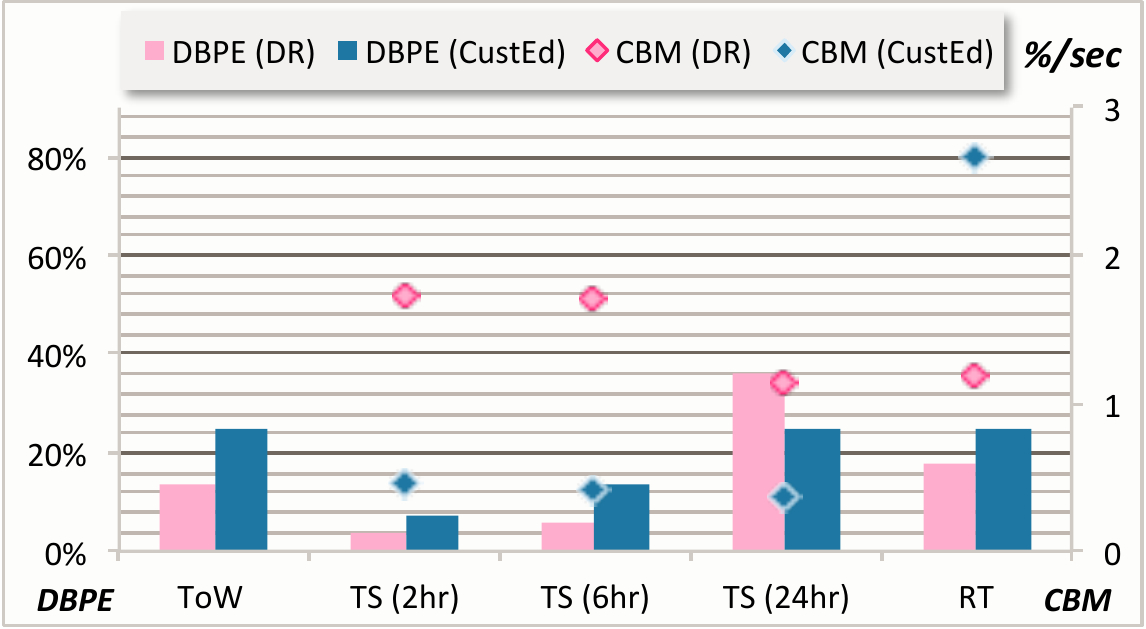}
   \caption{OFF}
        \label{fig:OHE15-DBPE}
 \end{subfigure}
 \begin{subfigure}[b]{\textwidth}
   \includegraphics[width=\textwidth] {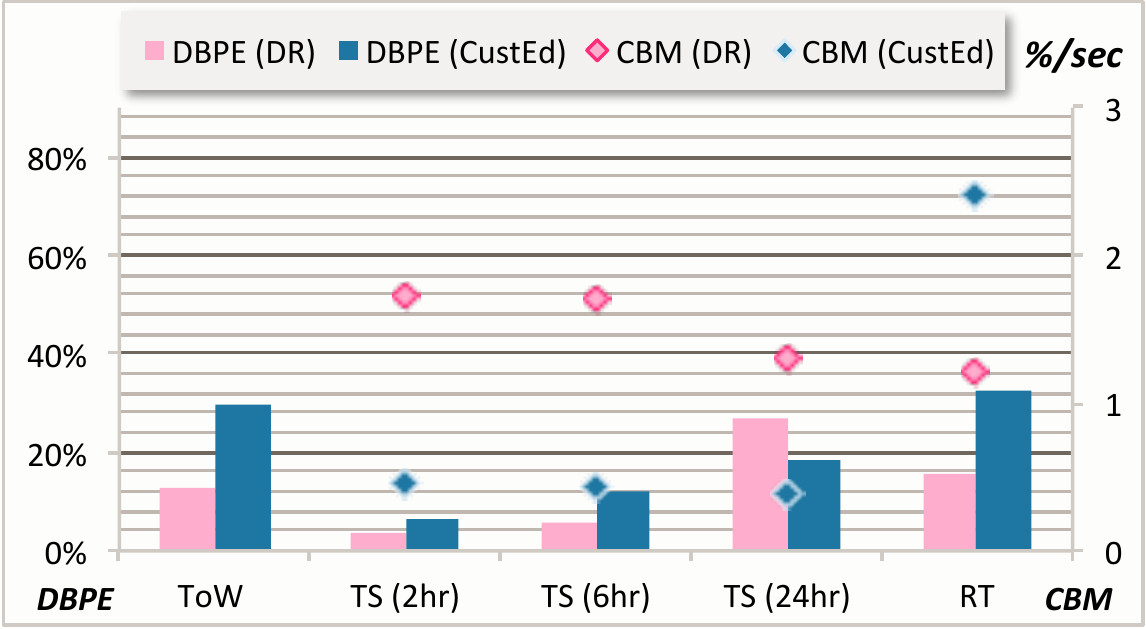}
   \caption{ACD}
        \label{fig:KAP15-DBPE}
 \end{subfigure}
\vspace{0.0in}
\caption{\textbf{\emph{DBPE}}, primary Y-axis, and \textbf{\emph{CBM}}, secondary Y-axis, for
  \textbf{\emph{15-min predictions}} for Demand Response and Customer Education. Customer Ed. is not
  relevant for campus. Lower DBPE and higher CBM are desirable, and provided by TS (2hr) and TS (6hr) for DR.}
\end{minipage}

\end{figure*}

Fig. \ref{fig:C24-DBPE} shows RT (and TS) to perform better than the DoW baseline on DBPE ($6.87\%$
vs. $7.56\%$, consistent with MAPE). The RT model's reliability is also higher than DoW's
(Fig. \ref{fig:C24-REL}), with RT providing errors smaller than the threshold $60.87\%$ of the time
-- a probabilistic measure for the planners to use. When we consider the total compute cost for
training and running the model (Table~\ref{table:cost}), RT is trained once a year and used
six times, with a negligible compute cost of $103$~msec and a high CBM of $900\%/sec$ 
(Fig.~\ref{fig:C24-DBPE}). These make RT a better qualitative and cost-effective model for long
term campus planning.


\textbf{Buildings:}
Buildings being upgraded for \add{sustainability and} energy efficiency favor over-prediction of
consumption to ensure \add{an} aggressive
reduction of carbon footprint\del{ without incurring huge expenses}. Reflecting this, we set $\alpha =
0.5$ and $\beta = 1.5$ for DBPE. A higher error tolerance than campus is acceptable, at
$e_{t}=15\%$. Cost parameters and measure values are \add{the} same as campus.

DBPE reflects a truer measure of error for the application and we see that it is smaller than MAPE
across all models and buildings 
(Fig \ref{fig:MHP24-DBPE}-\ref{fig:KAP24-DBPE}). Investigating the data reveals that the
\textit{average kWh for \add{the} training period was higher than that for the test period}, leading
to over-predictions. \add{Here,} the models' inclination to over-predict works in their
favor.\del{ in this case.} 
While RT is uniformly better than DoW on DBPE, 
it is less reliable for RES and ACD (Figs.~\ref{fig:FLT24-REL} \& \ref{fig:KAP24-REL}), even falling
below $0\%$, indicating \del{less frequent predictions below the error threshold than
  above.}\add{that predictions go over the error threshold more often than below the threshold.}

REL unlike DBPE treats over- and under-predictions
similarly. 
While the baseline has\del{ comparatively} fewer errors above the threshold, their magnitudes are much
higher, causing DBPE (an average) to rise \del{while}\add{for smaller} REL\del{ is smaller}.  
The costs for RT are minimal like for campus and their CBMs \del{comparable}\add{similar}. So the
model \add{of} choice depends
on if the predictions need to be below the threshold more often (DoW) or if the
biased-errors are lower (RT). \add{Particularly, for OFF, REL (Fig.~\ref{fig:OHE24-REL}) shows
  RT is best for Planning even as DoW was the better model based on CVRMSE
  (Fig.~\ref{fig:OHE24-CV-MAPE}). Similarly, for ACD, REL (Fig.~\ref{fig:KAP24-REL}) recommends DoW
  for Planning even as CVRMSE (Fig.~\ref{fig:KAP24-CV-MAPE}) suggests RT and MAPE
  (Fig.~\ref{fig:KAP24-CV-MAPE}) suggests TS (1wk). \emph{These highlight the value of defining
  application-specific performance measures like REL for meaningful model selection.}} 

\begin{figure*}
        \centering
        \begin{subfigure}[b]{0.3\textwidth}
                \centering
                \includegraphics[width=\textwidth]{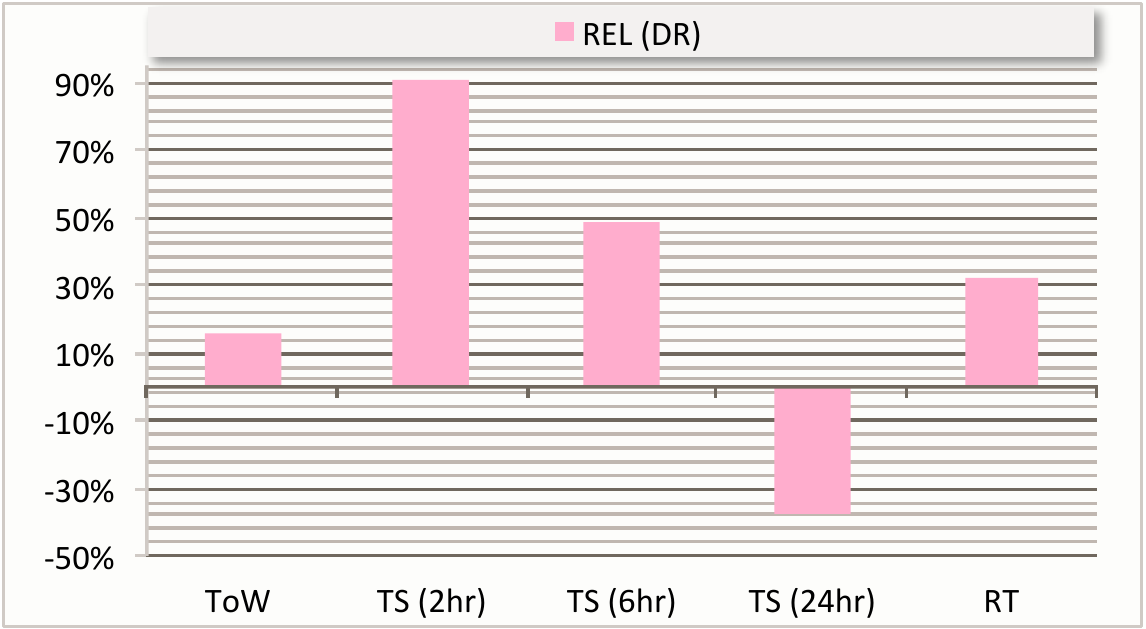}
                \caption{Campus}
                \label{fig:C15-REL}
        \end{subfigure}%
        ~ 
        \begin{subfigure}[b]{0.3\textwidth}
                \centering
                \includegraphics[width=\textwidth]{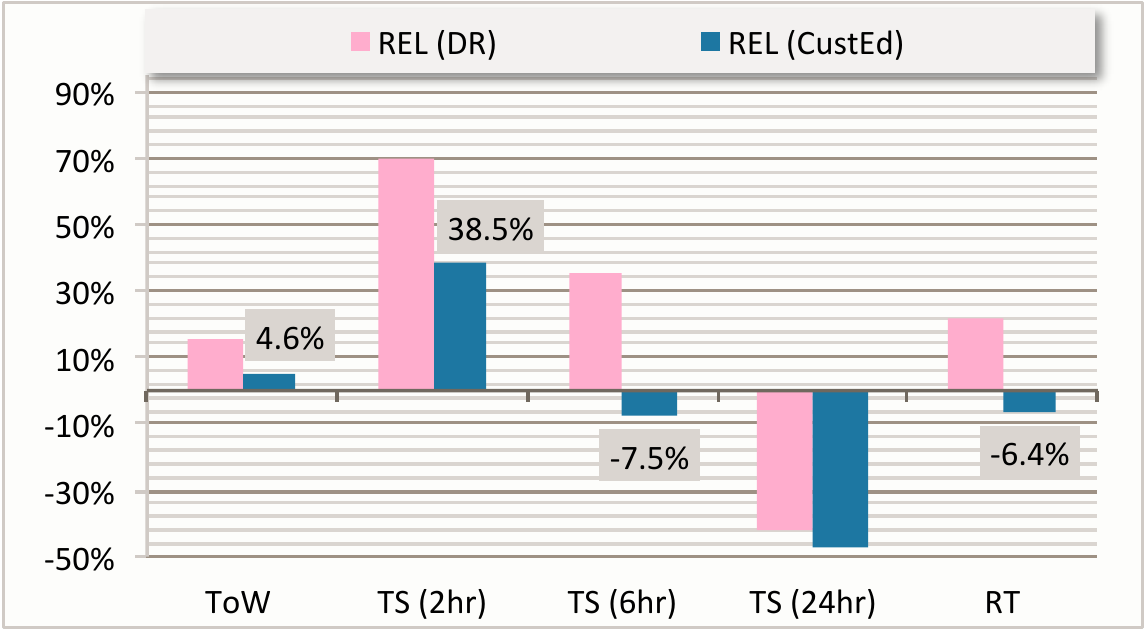}
                \caption{DPT}
                \label{fig:MHP15-REL}
        \end{subfigure}
        ~ 
        \begin{subfigure}[b]{0.3\textwidth}
                \centering
                \includegraphics[width=\textwidth]{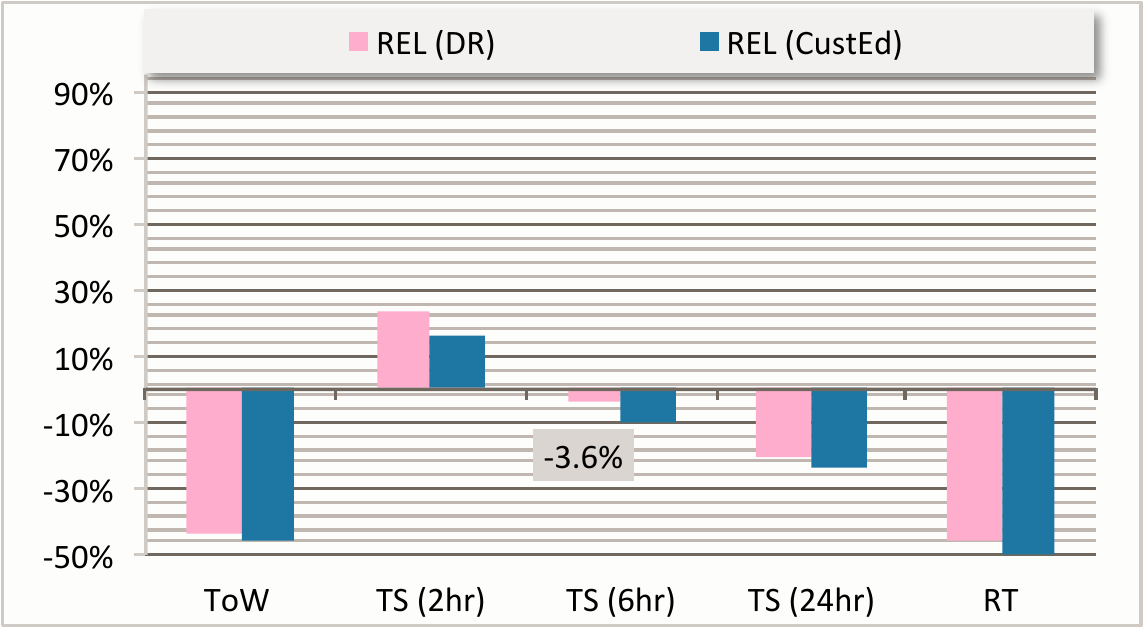}
                \caption{RES}
                \label{fig:FLT15-REL}
        \end{subfigure}
		\begin{subfigure}[b]{0.3\textwidth}
                \centering
                \includegraphics[width=\textwidth]{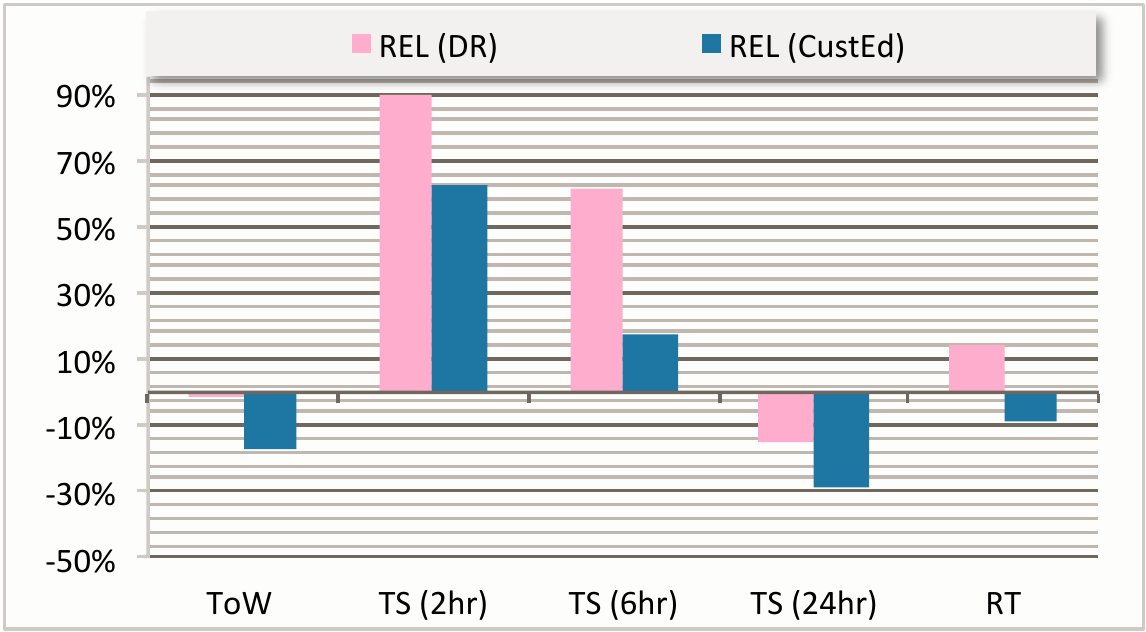}
                \caption{OFF}
                \label{fig:OHE15-REL}
        \end{subfigure}
		\begin{subfigure}[b]{0.3\textwidth}
                \centering
                \includegraphics[width=\textwidth]{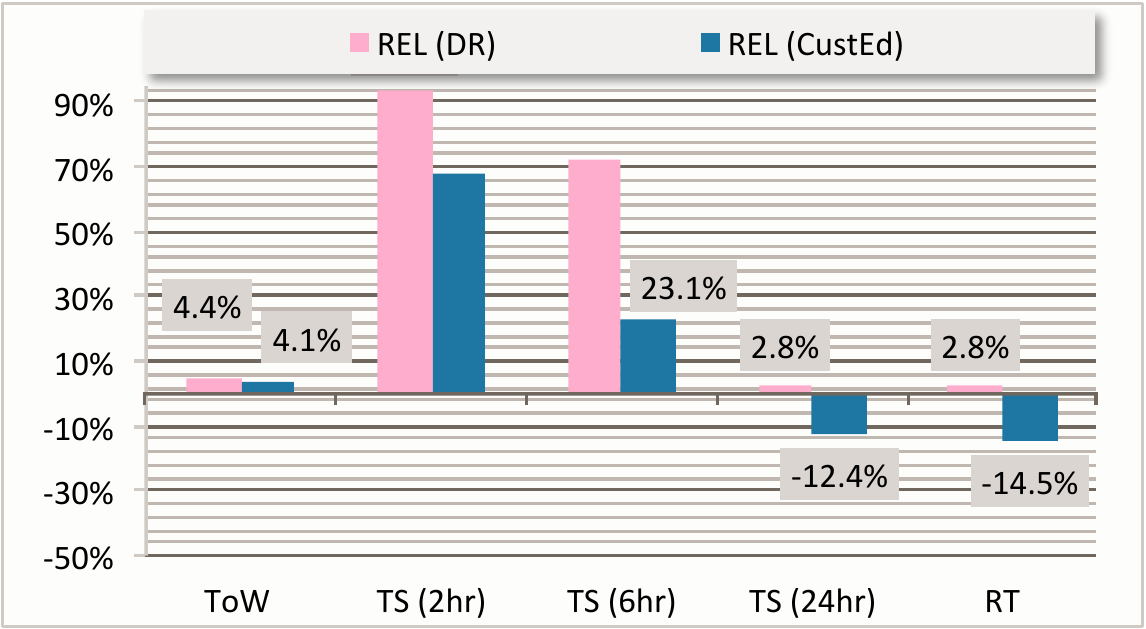}
                \caption{ACD}
                \label{fig:KAP15-REL}
        \end{subfigure}        
        \caption{\textbf{\emph{REL}} values for \textbf{\emph{15-min predictions}} for Demand Response and
          Customer Education. Higher is better.}
        \label{fig:REL-15}
\end{figure*}
\vspace{-0.1in}
\subsection{Customer Education}
This application uses 24-hour and 15-min predictions at the building-level made during the
daytime (6AM-10PM), and provides them to residents/occupants for monthly budgeting and daily energy conservation.

\textbf{24-hour predictions:} 
24-hour predictions impact monthly power budgets, and over-predictions are better to avoid 
slippage. We pick $\alpha = 0.75$ and $\beta = 1.25$ for DBPE and an error tolerance
$e_{t}=15\%$ for REL. We use a 4-week prediction duration for costing with one 24-hour prediction
done each day by RT and TS. RT is trained once in this \del{duration}\add{period}. We report TCC (Table~\ref{table:cost}),
DBPE \& CBM 
(Figs. \ref{fig:MHP24-DBPE}-\ref{fig:KAP24-DBPE}), and REL 
(Figs. \ref{fig:MHP24-REL}-\ref{fig:KAP24-REL}).

Like for Planning that preferred over-predictions, the DBPE here is smaller than MAPE for all
models, and it is mostly smaller for RT and TS models than DoW. \emph{But for a building like ACD, while
one may have picked TS (1wk) based on the application-independent MAPE measure
(Fig.~\ref{fig:KAP24-CV-MAPE}), both RT and DoW are better for Customer Education on DBPE
\add{(Fig.~\ref{fig:KAP24-DBPE})}.} \add{Similarly, for both DPT and RES, TS (1wk) was the best
option based on MAPE (Figs.~\ref{fig:MHP24-CV-MAPE}, \ref{fig:FLT24-CV-MAPE}) as well as on DBPE for
Customer Education (Figs.~\ref{fig:MHP24-DBPE}, \ref{fig:FLT24-DBPE}). However, for a different
application, such as Planning, RT is the recommended model based on DBPE
(Figs.~\ref{fig:MHP24-DBPE}, \ref{fig:FLT24-DBPE}). This highlights how a measure that is tailored
for a specific application by setting tunable parameters can guide the effective choosing of models for it.} 

When considering reliability, REL for RT is marginally (DPT) or significantly (OFF) better than
DoW even as the application-independent RIM showed RT to be worse or as bad as
DoW respectively -- yet another benefit of measures customized for the application. RT also
equals or out-performs TS (1wk) on both DBPE and REL on all buildings but RES. The TCC cost for TS
while being $\sim20\times$ more than RT is still small given the one month duration. This also
reflects in the CBM being much lower for TS.




\textbf{15-min predictions:} 
This \del{help}\add{application} engages customers by giving periodic \del{predictions}\add{forecasts} during the day to encourage
efficiency. Over-predicting often or more frequent errors will mitigate \add{a} customer's interest. So we
set $\alpha = 1.5$ and $\beta = 0.5$ for DBPE, and we have a lower error tolerance at $e_t=10\%$ for
REL. Prediction duration is 4-weeks for cost parameters, with 8 uses per day at 2~hour horizons. RT
is trained once.

For all buildings, both DBPE and REL rank TS (2hr) as the best model 
(Figs.~\ref{fig:MHP15-DBPE}-\ref{fig:KAP15-DBPE}) \&
(Figs.~\ref{fig:MHP15-REL}-\ref{fig:KAP15-REL}). These reaffirm the effectiveness of TS \del{in}\add{for} short-term
predictions\del{ and its
emphasis of near-term data over historic averages that may shift}. For many models, the (daytime)
DBPE for this application is higher than the (all-day) MAPE due to higher variations in the
day. However, TS (2hr) bucks this trend for RES, OFF and ACD. RT is worse than even ToW on reliability, with REL
below 0\% for all buildings. For RES, all models but\del{ for} TS (2hr) have REL below 0\%. So
qualitatively, TS (2hr) is by far a better model. However, on costs (Table~\ref{table:cost}), TS
has TCC~$\approx 209$~secs. This may not seem much but when used for 10,000's of buildings in a
utility\del{ area}, it can be punitive. At large scales, CBM 
(Figs.~\ref{fig:MHP15-DBPE}-\ref{fig:KAP15-DBPE}) may offer a \add{better} trade-off and
suggest RT for DPT and OFF.  

\begin{table}[!t]
\renewcommand{\arraystretch}{1.0}

\caption{\textbf{Application Specific Cost Parameters \& TCC}. $\tau$ is the trainings per duration,
  and $\pi$ is the model usage with a prediction horizon per duration.}
\centering  
\begin{tabular}{l c r|r} 

\hline\hline                  
Application & Trainings, $\tau$ & Uses, $\pi$ (horizon) & TCC (millisec) \\ 
\hline
\multicolumn{3}{l|}{\textbf{Planning} (\textit{duration = 1 year})} & \\
24-hour RT & 1 & 6 (2mo) & 103 \\
\hline
\multicolumn{3}{l|}{\textbf{Customer Education} (\textit{duration = 4 weeks})} & \\
24-hour RT & 1 & 28 (1dy) & 139\\
24-hour TS & - & 28 (1dy) & 2,845\\
15-min RT &  1 & 8$\cdot$28 (2hr) & 28,103 \\
15-min TS & - & 8$\cdot$28 (2hr) & 2,09,037\\
\hline
\multicolumn{3}{l|}{\textbf{Demand Response} (\textit{duration = 4 weeks})} & \\
15-min RT & 4 & 5$\cdot$3 (6hr) & 69,824\\
15-min TS & - & 4$\cdot$5$\cdot$3 (6hr) & 55,992\\
\hline
\end{tabular}
\label{table:cost} 
\end{table} 
 
\subsection{Demand Response}
DR uses 15-min predictions to detect peak usage and preemptively correct them to prevent grid
instability. Hence, over-predictions are favored \del{over}\add{than} under to avoid missing peaks, and we set
$\alpha = 0.5$ and $\beta = 1.5$ for DBPE. The campus is a large customer with tighter
requirements of error threshold at $e_{t}=5\%$ for REL, while individual buildings with lower impact
are allowed a wider error margin of $e_t=10\%$. Prediction duration is 4 weeks for cost
parameters, with the models used thrice a weekday -- before, at the start
and during the 1-5PM period, and RT trained weekly.

DBPE is uniformly smaller than MAPE for the campus and buildings 
(Figs.~\ref{fig:C15-DBPE}-\ref{fig:KAP15-DBPE}), sometimes even halving the
errors. Thus the 4~hour DR periods in the weekdays are more (over-)predictable than all-day predictions. TS
(2hr) has significantly better DBPE than other models, with even TS (6hr) out-performing RT and
ToW. For campus, RT is better than DoW, in part due to using temperature
features that have a cumulative impact on 
energy use during midday. 

We see TS (2hr) gives a high REL of $91\%$ for campus 
(Fig.~\ref{fig:C15-REL}) and is the
only model with positive REL for RES. Also, TS (6hr)
and RT prove to be more reliable for DR in campus and DPT than their poorer showing in the RIM and VAB
independent measures (Fig.~\ref{fig:RIM-VAB-15}), making them competitive candidates. However, RT suffers in reliable
predictions for other buildings, with lower or negative REL (Figs.~\ref{fig:FLT15-REL}-\ref{fig:KAP15-REL}) while TS (2hr) and
(6hr) continue to perform reliably. 



Cost-wise, we see RT and TS models are comparable on TCC (Table~\ref{table:cost}). For once, RT takes
longer than TS due to the more aggressive retraining (every week), preferred for
critical DR operations. But when seen through the CBM measure, all TS models beat RT for all
cases but one (TS (24hr) on DPT). Thus, the TS (2hr) and TS (6hr) are the best for DR on all measures.

\section{Conclusion}
\label{conclusion}
\add{The key consideration in evaluating a prediction model by an end-user is its performance for
  the task at hand. Traditionally, accuracy measures have been used as the sole measure of
  prediction quality.} In this \del{paper}\add{article}, we examine \add{the value of} holistic performance measures along the dimensions of scale independence,
reliability and cost. In evaluating them for consumption prediction in Smart Grids, we see that
\textit{scale independence} ensures that performance can be compared across models and applications
and for different customers; \textit{reliability }evaluates a model's consistency of performance
with respect to baseline models\del{ used in practice}; while \textit{cost }is a key consideration when
deploying models at large scale for real world applications. We use \add{existing} scale-independent measures,
\textit{CVRMSE} and \textit{MAPE}, while extending and proposing four additional measures,
\textit{RIM} and \textit{VAB} for measuring reliability; and \textit{CD} and \textit{CC} for data
and compute costs. 

Further, our novel application-dependent measures can be customized by domain experts for
meaningful model evaluation for applications of interest. These measures include \textit{DBPE} for
scale independence, \textit{REL} for reliability, and \textit{TCC} and \textit{CBM} for
cost. \add{The value of these measures for scenario-specific model selection were empirically
  demonstrated using}\del{The}
three Smart Grid applications that anchored our analysis\add{ even as they are generalizable to other domains}. Through cross correlation analysis, we found that only MAPE and CVRMSE show absolute correlation $>0.9$, indicating that all measures are individually useful. \del{helped}\add{Our results} demonstrate
the \add{valuable} insight\add{s} that \del{could}\add{can}
be gleaned on models' behavior \add{using holistic measures.} \add{These help} to improve their performance, and provide an understanding of the
predictions' real impact in a comprehensive yet accessible manner. \del{Our measures are
generalizable to other domains}\add{As such, they offer} a common frame of reference for
\add{model evaluation by} future researchers and
practitioners.
 

%

\appendices

\section*{Acknowledgments}


This material is based upon work supported by the United States Department of Energy under Award
Number DEOE0000192, and the Los Angeles Department of Water and Power (LA DWP). The views and
opinions of authors expressed herein do not necessarily state or reflect those of the United States
Government or any agency thereof, the LA DWP, nor any of their employees. 
\ifCLASSOPTIONcaptionsoff
  \newpage
\fi



%

\bibliographystyle{abbrv}
\bibliography{saimaref}



%

\begin{IEEEbiographynophoto}{Saima Aman}
is a PhD student in Computer Science at the Univ of Southern California and a Research Assistant in the Smart Grid project at the Center for Energy Informatics. Her research interests are in the areas of Data Mining and Artificial Intelligence. She has a Master's in Computer Science from the University of Ottawa. 
\end{IEEEbiographynophoto}
\vfill
\vspace{-0.4in}\begin{IEEEbiographynophoto}{Yogesh Simmhan} is an Assistant Professor in the Indian Institute of Science, and previously a faculty at the University of Southern California and a postdoc at Microsoft Research. His research is on scalable programming models for Big Data applications on distributed systems like Clouds. He has a PhD in Computer Science from Indiana University. Senior member of IEEE and ACM.
\end{IEEEbiographynophoto}
\vfill
\vspace{-0.4in}\begin{IEEEbiographynophoto}{Viktor K. Prasanna}
is the Powell Chair in Engineering, Professor of Electrical Engineering
and Computer Science, and Director of the Center for Energy Informatics at the
Univ of Southern California. His research interests include HPC,
Reconfigurable Computing, and Embedded Systems. He received his MS from the School of Automation, Indian Institute of Science and PhD in
Computer Science from Penn State. He is a Fellow of IEEE, ACM and AAAS.
\end{IEEEbiographynophoto}
\vfill
\end{document}